\documentclass[lettersize,journal]{IEEEtran}
\usepackage{amsmath,amsfonts}
\usepackage{algorithmic}
\usepackage{algorithm}
\usepackage{array}
\usepackage[caption=false,font=normalsize,labelfont=sf,textfont=sf]{subfig}
\usepackage{textcomp}
\usepackage{stfloats}
\usepackage{url}
\usepackage{verbatim}
\usepackage{graphicx}
\usepackage{cite}
\usepackage{booktabs}     
\usepackage{tabularx}    
\usepackage{array}       
\usepackage{multirow}   

\usepackage[table]{xcolor} 
\usepackage[hidelinks]{hyperref}
\begin{document}

\title{Drift-Aware Online Dynamic Learning for Nonstationary Multivariate Time Series: Application to Sintering Quality Prediction}

\author{Yumeng~Zhao,
        Shengxiang~Yang,~\IEEEmembership{Fellow,~IEEE,}
        and~Xianpeng~Wang,~\IEEEmembership{Senior~Member,~IEEE}%
\thanks{This work was supported by the China Scholarship Council (CSC), No.~202406080121, and in part by the National Natural Science Foundation of China (62473086). (Corresponding authors: Shengxiang Yang and Xianpeng Wang.)}%
\thanks{Yumeng Zhao is with the the National Frontiers Science Center for Industrial Intelligence and Systems Optimization, Northeastern University, Shenyang, China, and also with De Montfort University, Leicester, U.K. (e-mail: \texttt{yumengzhao0606@outlook.com}).}%
\thanks{Shengxiang Yang is with the Innovative Artificial Intelligence Cluster, Digital Future Institute, De Montfort University, Leicester, U.K. (e-mail: \texttt{syang@dmu.ac.uk}).}%
\thanks{Xianpeng Wang is with the Key Laboratory of Data Analytics and Optimization for Smart Industry (Northeastern University), Ministry of Education, Shenyang, China (e-mail: \texttt{wangxianpeng@ise.neu.edu.cn}).}%
}

% ===== Page headers (replace the IEEEtran sample header) =====
\markboth{IEEE Transactions on Neural Networks and Learning Systems}%
{Zhao \MakeLowercase{\textit{et al.}}: Drift-Aware Online Dynamic Learning for Nonstationary Multivariate Time Series}

% ===== Submission stage: do NOT include the template pubid line =====

\maketitle

\begin{abstract}

Accurate prediction of nonstationary multivariate time series remains a critical challenge in complex industrial systems such as iron ore sintering.
In practice, pronounced concept drift compounded by significant label verification latency rapidly degrades the performance of offline-trained models.
Existing methods based on static architectures or passive update strategies struggle to simultaneously extract multi-scale spatiotemporal features and overcome the stability-plasticity dilemma without immediate supervision.
To address these limitations, a Drift-Aware Multi-Scale Dynamic Learning (DA-MSDL) framework is proposed to maintain robust multi-output predictive performance via online adaptive mechanisms on nonstationary data streams.
The framework employs a multi-scale bi-branch convolutional network as its backbone to disentangle local fluctuations from long-term trends, thereby enhancing representational capacity for complex dynamic patterns.
To circumvent the label latency bottleneck, DA-MSDL leverages Maximum Mean Discrepancy (MMD) for unsupervised drift detection. By quantifying online statistical deviations in feature distributions, DA-MSDL proactively triggers model adaptation prior to inference.
Furthermore, a drift-severity-guided hierarchical fine-tuning strategy is developed. Supported by prioritized experience replay from a dynamic memory queue, this approach achieves rapid distribution alignment while effectively mitigating catastrophic forgetting.
Long-horizon experiments on real-world industrial sintering data and a public benchmark dataset demonstrate that DA-MSDL consistently outperforms representative baselines under severe concept drift.
Exhibiting strong cross-domain generalization and predictive stability, the proposed framework provides an effective online dynamic learning paradigm for quality monitoring in nonstationary environments.
\end{abstract}


\section{Introduction}

\IEEEPARstart{T}{he} modeling and forecasting of nonstationary multivariate time series have emerged as fundamental pillars across diverse domains, including industrial process control, energy management, and financial risk mitigation~\cite{Lu2018Learning}.
While standard forecasting paradigms typically map historical feature sequences to target variables, most existing approaches implicitly rely on the stationarity assumption.
They presuppose an invariant conditional distribution $p_t(y|x)$ over time, allowing for the acquisition of stable models via one-time offline training~\cite{Gielczyk2024Evaluation}.
Unfortunately, this prerequisite is frequently violated in complex real-world scenarios. In large-scale industrial processes, for instance, fluctuating raw material compositions, equipment degradation, and evolving operational strategies inevitably drive the continuous evolution of data distributions~\cite{Hu2024Dynamic}.
Consequently, static models trained under the independent and identically distributed (i.i.d.) assumption suffer severe performance degradation—such as surging prediction errors, prolonged recovery durations, and systematic biases—when encountering distributional shifts, commonly known as concept drift~\cite{Yan2025BTPNet}.
Therefore, developing a learning system capable of long-term continuous adaptation—without relying on frequent and computationally expensive full retraining—remains a pivotal challenge in nonstationary time series forecasting.

These challenges are epitomized by the iron ore sintering process, a complex dynamic system characterized by intricate heat transfer, combustion, and phase-change reactions, exhibiting pronounced nonlinearity and multi-scale time-varying characteristics~\cite{Xue2024Phase-Field, Olevsky2006Multi-Scale}.
 In this context, slow-evolving trends and abrupt operational fluctuations intertwine, driving the continuous evolution of features, labels, and the underlying conditional distribution $p_t(y|x)$~\cite{Hu2024Dynamic}.
Moreover, distinct from conventional academic benchmark datasets, quality indicators in industrial scenarios typically rely on offline laboratory assays, inevitably introducing significant label verification latency~\cite{9369974Hu,10167823Yao}.
By severing the instantaneous error-driven feedback loop, this extreme verification latency transforms the online adaptation process into an ill-posed optimization problem, since the true empirical risk cannot be evaluated in real time~\cite{Hu2024Soft}.
To this end, the system leverages unsupervised statistical discrepancy measures to approximate joint distribution shifts, thereby imposing prior information-theoretic constraints on the parameter search space.
Therefore, the core bottleneck for long-term predictive reliability lies in executing multi-scale online updates without immediate feedback. Specifically, the learning system must maintain a principled balance the retention of historical knowledge with the rapid absorption of new distribution patterns.

Existing methods for nonstationary time series forecasting generally fall into three categories. 
The first category enhances static representation capabilities using sophisticated operators such as multi-scale convolutions or attention mechanisms~\cite{LaraBenitez2021TCN, Xie2024EnvFormer}. 
The second category develops drift detection mechanisms that trigger model retraining when prediction errors increase significantly~\cite{Gielczyk2024Evaluation}. 
The third category explores parameter fine-tuning or knowledge transfer across different data distributions~\cite{Yang2023Multisource}.
Despite these advancements, most methods rely on a reactive ``predict-evaluate-update'' feedback paradigm. 
This paradigm assumes that supervisory signals (labels) are immediately available to compute real-time errors and guide model updates. 
However, in industrial processes such as sintering, severe label latency severs this error-driven feedback loop. As a result, models can only compensate after a long-delayed performance drop, which is inadequate for abrupt operational shifts~\cite{Yan2022DSTED}.
Furthermore, existing multi-scale feature extractors and online update mechanisms are typically designed as isolated modules. 
This decoupling causes complex architectures to be treated as ``black boxes'' during coarse-grained fine-tuning. Such an approach is computationally expensive, highly susceptible to industrial noise, and makes it difficult to achieve a controlled trade-off between accuracy improvement and update cost.

To achieve reliable online quality prediction under severe concept drift and significant label latency, we propose a Drift-Aware Multi-Scale Dynamic Learning (DA-MSDL) framework for nonstationary multivariate time series forecasting.
The framework leverages a multi-scale bi-branch convolutional network as its backbone to decouple complex spatiotemporal dependencies, and integrates an unsupervised Maximum Mean Discrepancy (MMD)-based detection mechanism to identify distribution shifts in the absence of real-time labels.
Unlike conventional reactive update paradigms, DA-MSDL adopts a proactive ``detect-fine-tune-predict'' pipeline. Specifically, a drift-severity-guided transfer-based hierarchical fine-tuning strategy enables the model to adjust its parameters before inference, thereby compensating for drift under delayed supervision.
Extensive experiments on long-horizon real-world industrial sintering data and a public industrial benchmark dataset validate the effectiveness of DA-MSDL for nonstationary industrial data streams.
The major contributions of this study are summarized as follows:
\begin{itemize}
    \item[1)] A unified DA-MSDL framework is proposed to integrate decoupled multi-scale feature representation with unsupervised drift detection for nonstationary prediction in complex industrial processes.
    \item[2)] A proactive transfer-based adaptation mechanism is developed to perform drift-severity-guided hierarchical fine-tuning before inference, thereby reducing dependence on immediate label feedback.
    \item[3)] A prioritized replay mechanism based on a dynamic memory queue is introduced to balance stability and plasticity, enabling rapid adaptation to evolving distributions while mitigating catastrophic forgetting.
    \item[4)] Extensive long-horizon validation on real-world industrial sintering data and a public industrial benchmark dataset demonstrates the superior accuracy and robustness of DA-MSDL for quality monitoring in nonstationary industrial environments.
\end{itemize}

The remainder of this paper is organized as follows.
Section~\ref{Section2} reviews the related works on nonstationary time series forecasting and online learning.
Section~\ref{Section3} elaborates on the proposed DA-MSDL framework and its key modules.
Section~\ref{Section4} presents the experimental results and provides a detailed performance analysis.
Finally, Section~\ref{Section5} concludes this paper.

\section{Problem Statement and Related Work}
\label{Section2}

\subsection{Problem Statement}

\begin{figure}[t]
\centering
\includegraphics[width=\linewidth]{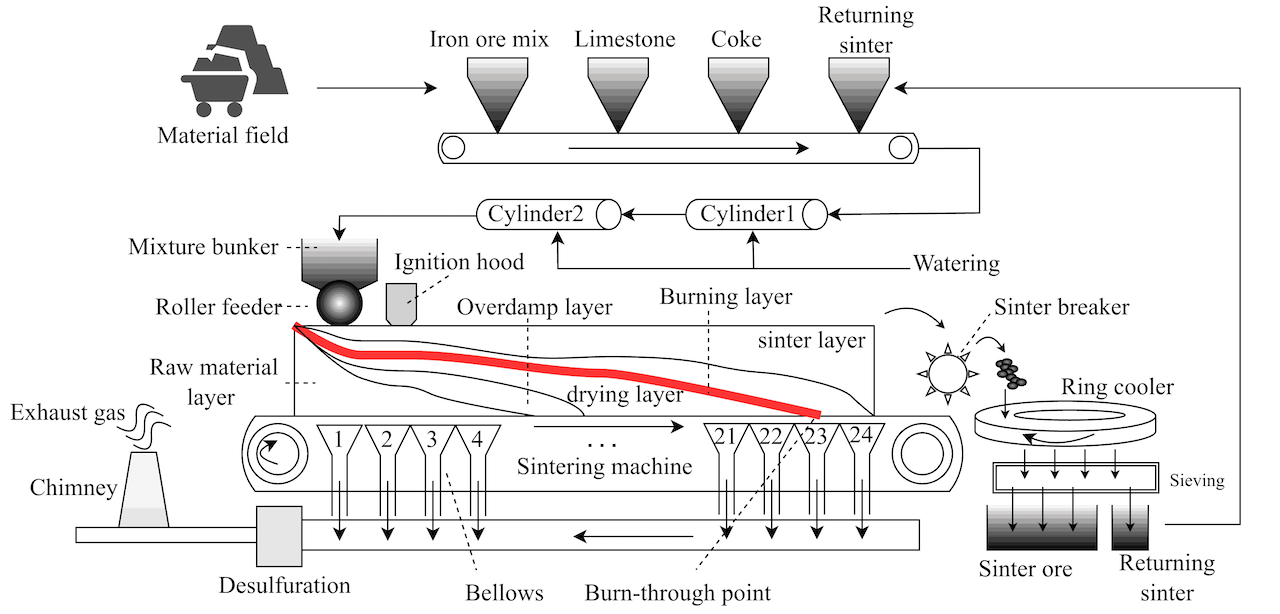}
\caption{Schematic diagram of the sintering process.}
\label{fig:sinter_process}
\end{figure}

Iron ore sintering is a typical belt-type agglomeration process that transforms blended fine ores into sinter under high-temperature conditions.
The sintering process involves a complete sequence of physicochemical stages, including material granulation, ignition, suction sintering, cooling, and crushing, as illustrated in Fig.~\ref{fig:sinter_process}.
It exhibits strong nonlinearity, multi-scale temporal variability, and complex couplings among process variables, making online quality forecasting susceptible to persistent distribution shifts during long-horizon deployment.
Meanwhile, frequent changes in raw materials and operating strategies further induce operational drift, causing the underlying data distribution to evolve over time.
Moreover, key quality indices are typically obtained via end-point offline laboratory assays, so the ground-truth labels cannot be returned in real time and exhibit verification latency.

For iron ore sintering, a representative industrial dynamic system, we formulate quality index prediction as a nonstationary multivariate time series forecasting problem. 
Let $\mathcal{S}={(\mathbf{x}_t,\mathbf{y}_t)},{t=1,\dots,T}$ denote a continuously arriving production data stream, where each observation at time $t$ consists of an $F$-dimensional process feature vector $\mathbf{x}_t\in\mathbb{R}^F$ and a $K$-dimensional quality target vector $\mathbf{y}_t\in\mathbb{R}^K$.

The objective is to learn a mapping function $f_{\theta_t}$ that evolves with the data distribution over time, where $\theta_t$ denotes the vector of model parameters at time $t$.
Given a historical observation window of length $L$, the predictive output at time $t$ is defined as:
\begin{equation}
    \hat{\mathbf{y}}_t = f_{\theta_t}(\mathbf{X}_{t-L+1:t}), \quad t = L, \dots, T
    \label{eq:evolutionary_model}
\end{equation}
where $\mathbf{X}_{t-L+1:t}\in\mathbb{R}^{F\times L}$ denotes the input matrix that captures local spatiotemporal dependencies.

Due to continuous operational drift in sintering, we explicitly assume that the underlying joint distribution $P_t(\mathbf{x},\mathbf{y})$ evolves over time.
This nonstationarity causes a static model $f_{\theta_0}$ trained on a fixed distribution to suffer significant performance degradation during long-term operation, i.e.,
\begin{equation}
    \exists t, k: P_t(\mathbf{x}, \mathbf{y}) \neq P_{t+k}(\mathbf{x}, \mathbf{y})
    \label{eq:nonstationary_evolution}
\end{equation}
where $k\in\mathbb{Z}^{+}$ denotes a positive time lag over which the operating condition evolves.

Furthermore, due to industrial offline assay cycles, the system faces severe label verification latency, where the ground-truth label $\mathbf{y}_t$ at time $t$ becomes available only after $\tau$ time steps.
Consequently, when adaptive fine-tuning is performed at time $t$, the available supervisory signals are restricted to the lagged sequence:
\begin{equation}
    \mathcal{D}_{obs} = \{(\mathbf{x}_i, \mathbf{y}_i)\}_{i=1}^{t-\tau}
    \label{eq:latency_constraint}
\end{equation}

Under the aforementioned streaming constraints and label verification latency, our goal is to develop a proactive online adaptation paradigm.
Instead of one-time offline optimization, we formulate the learning process as a sequential decision problem.
At each time step $t$, the objective is to update the model parameters locally so as to minimize instantaneous predictive risk while maintaining adaptation stability.
The global performance over the entire operating horizon is evaluated by the cumulative normalized predictive risk $\mathcal{J}$, defined as:
 \begin{equation}
    \min_{\{\theta_t\}_{t=L}^T} \mathcal{J} = \sum_{t=L}^{T} \left[ \mathcal{L}_{norm}\left( \hat{\mathbf{y}}_t, \mathbf{y}_t \right) + \lambda \Omega(\theta_t, \theta_{t-1}) \right]
    \label{eq:optimization_target}
\end{equation}
where $\mathcal{L}{norm}(\cdot)$ denotes the normalized forecasting loss, which serves as an empirical proxy for instantaneous risk at time $t$. $\Omega(\theta_t,\theta_{t-1})$ denotes a stability regularization term that encourages smooth parameter evolution and mitigates catastrophic forgetting, and $\lambda>0$ is a hyperparameter balancing forecasting accuracy and adaptation stability.

\subsection{Sintering Modeling and Forecasting}

The iron ore sintering process is a representative industrial system characterized by strong nonlinearity, strong coupling, and complex time-varying behavior~\cite{Xue2024Phase-Field}.
To describe physicochemical evolution in sintering, researchers initially developed mechanistic models based on conservation laws, such as phase-field simulations for microstructural evolution~\cite{Xue2024Phase-Field, Olevsky2006Multi-Scale}.
While these models provide valuable physical insights, their high computational complexity in solving heat and mass transfer equations makes it difficult to meet the real-time prediction requirements of industrial production~\cite{Gui2019Review}. 
To overcome these computational bottlenecks, data-driven methods were introduced. For example, Gao et al. used selective support vector regression (SVR) to predict sinter basicity~\cite{gao2016modeling}.
Furthermore, Wang et al. also employed Extreme Learning Machines (ELM) for rapid modeling in chemical composition prediction, due to their fast training speed~\cite{Wang2015ELM}.
However, these shallow architecture-based approaches typically rely on manual feature engineering and struggle to capture high-order nonlinear features embedded within raw signals~\cite{Yin2014Review}.

With the development of deep learning, Xie et al. utilized Long Short-Term Memory (LSTM) networks to capture long-range temporal dependencies inherent in the sintering process~\cite{Xie2021LSTM}. 
To address the frequency mismatch between high-frequency features and low-frequency labels in sintering data, several studies have adopted sophisticated sliding-window synchronization strategies~\cite{Liu2020Sync}. 
Subsequently, attention-based models were introduced to quantify the contributions of different process features to final quality indices through dynamic weight assignment~\cite{Zhang2022Attention}.
To further model complex thermal interactions and address multi-step forecasting challenges, advanced architectures such as Transformers have been adapted to sintering scenarios. For instance, EnvFormer, a decomposition-based Transformer, uses an iterative envelope decomposition module to improve burn-through point prediction~\cite{Xie2024EnvFormer}.
To capture the multi-scale dynamics of sintering, Dong and Yan demonstrated that multi-flow multi-scale convolutional networks can improve the representation of nonstationary signals~\cite{Dong2024Novel}.
Despite the superior performance of these deep models, most of them implicitly assume that the data distribution remains stationary throughout long-term operation~\cite{Gielczyk2024Evaluation}. 
However, frequent raw material adjustments and operating condition switching can induce significant distribution shifts, leading to rapid performance degradation of static models~\cite{Yan2025BTPNet}.
Consequently, under the constraint of delayed label supervision, achieving real-time drift awareness and online adaptation for nonstationary sintering remains challenging~\cite{Hu2024Dynamic}.

\subsection{Online Learning under Concept Drift and Latency}

Concept drift refers to the phenomenon in which the underlying distribution of a data stream changes over time, and it has become one of the central challenges in online learning systems~\cite{Lu2018Learning}.
Traditional drift detection methods primarily rely on model predictive performance. For instance, the Drift Detection Method (DDM) identifies distribution shifts by monitoring abrupt changes in classification or regression errors~\cite{Gielczyk2024Evaluation}.
Subsequently, statistical methods such as Adaptive Windowing (ADWIN) were proposed to capture drift signals by comparing data statistics across different sliding windows~\cite{Gielczyk2024Evaluation}.
Considering the cost of label acquisition, researchers have explored unsupervised drift monitoring using kernel methods such as Maximum Mean Discrepancy (MMD) to quantify statistical discrepancies in feature space~\cite{shao2022unsupervised}.
Verification latency is a challenging constraint in stream data mining, referring to the temporal gap between sample arrival and the availability of ground-truth labels~\cite{dyer2014compose}.
To mitigate model degradation caused by missing feedback, several studies have explored semi-supervised learning and self-labeling techniques to maintain model performance during unlabeled periods~\cite{triguero2015self}.
However, in highly nonstationary environments, pseudo-label accuracy is difficult to guarantee and may lead to severe error accumulation during online fine-tuning~\cite{Hu2024Soft}.
To compensate for delayed feedback, Souza et al. proposed a clustering-based classifier update framework (SCARGC) to estimate distribution trends in the absence of supervisory signals~\cite{souza2015scargc}. 
Although these methods perform reasonably well in classification tasks, their robustness remains insufficient in industrial regression prediction tasks characterized by continuous output spaces and complex multivariate coupling~\cite{Yin2014Review}. 
With the growing demand for online learning, researchers have begun exploring adaptive network architectures for data streams, including preliminary attempts to apply Neural Architecture Search (NAS) in streaming environments~\cite{Caldas2024Online}.
However, existing online NAS methods typically incur substantial computational overhead, making real-time architecture adaptation difficult under the millisecond-level response requirements of industrial production~\cite{elsken2019neural}.

After drift is detected, online learning frameworks typically employ incremental updates or retraining strategies to absorb knowledge from new distributions in real time~\cite{Lu2018Learning}.
Transfer learning techniques have also been introduced into online settings, where feature representations learned in the source domain are adapted to the shifted target domain through parameter fine-tuning~\cite{Yang2023Multisource}.
To improve fine-tuning efficiency, some studies have investigated hierarchical update mechanisms that dynamically freeze different layers of the model backbone according to drift severity, so as to balance computational overhead and adaptation speed~\cite{Hu2025Adaptive}.
However, most of the aforementioned methods are prone to catastrophic forgetting during online fine-tuning and implicitly assume the immediate availability of supervisory signals, thus often failing in real-world industrial environments characterized by label verification latency and high-intensity noise~\cite{kirkpatrick2017overcoming}.
During online parameter updates, the key challenge is to balance plasticity for integrating new knowledge with stability for preventing catastrophic interference with previously learned knowledge, i.e., the stability-plasticity dilemma~\cite{parisi2019continual}.
Incremental learning strategies have been introduced into online learning to mitigate catastrophic forgetting by maintaining a queue of historical samples~\cite{Hu2025Adaptive}. 
However, in industrial data streams with high-intensity noise and outliers, simple sample replay often causes models to overfit stochastic fluctuations rather than true operational drift~\cite{shao2022unsupervised}. 
Considering the multi-scale dynamics in industrial processes, existing single-scale online learning models struggle to simultaneously capture long-period evolution and instantaneous abrupt disturbances~\cite{LaraBenitez2021TCN}. 
Most existing studies still follow a ``passive retraining after performance decline'' logic and lack mechanisms for proactive adjustment based on feature distribution information~\cite{Yan2022DSTED}.
In other dynamic decision-making problems, prediction-driven adaptation has been explored to proactively respond to environmental changes by modeling spatial--temporal trajectories, thereby accelerating updates under dynamic environments~\cite{10440544Zhao}.
To address these limitations, it is imperative to develop a proactive dynamic learning framework that integrates unsupervised drift detection with hierarchical model fine-tuning, thereby ensuring robust prediction under nonstationary and label-scarce conditions.

\section{Drift-Aware Multi-Scale Dynamic Learning Framework}
\label{Section3}

To address nonstationarity and label verification latency in sintering, this section presents the DA-MSDL framework for proactive online forecasting adaptation.
We then introduce the overall architecture and backbone, the construction and update of the Dynamic Memory Queue, the unsupervised MMD-based drift detection and decision logic, and finally the trend-aware hierarchical fine-tuning mechanism that integrates drift-driven transfer adaptation with stable error calibration.

\subsection{Framework Overview}
\label{sec:overview}

The representational foundation of the DA-MSDL framework is provided by the Multi-Scale Branched Convolutional Neural Network (MS-BCNN). 
As the backbone architecture has been described previously~\cite{mypaper}, 
we provide a brief overview here and include full architectural details and hyperparameters in the Supplementary Material.
MS-BCNN performs multi-scale feature disentanglement through parallel short- and long-kernel convolution branches, which capture local fluctuations and long-term trends in complex industrial signals.
This design enables the model to learn robust multi-scale representations under drastic operational fluctuations, thereby providing stable representational support for subsequent online adaptation.

\begin{figure*}[!t]
\centering
\includegraphics[width=\textwidth]{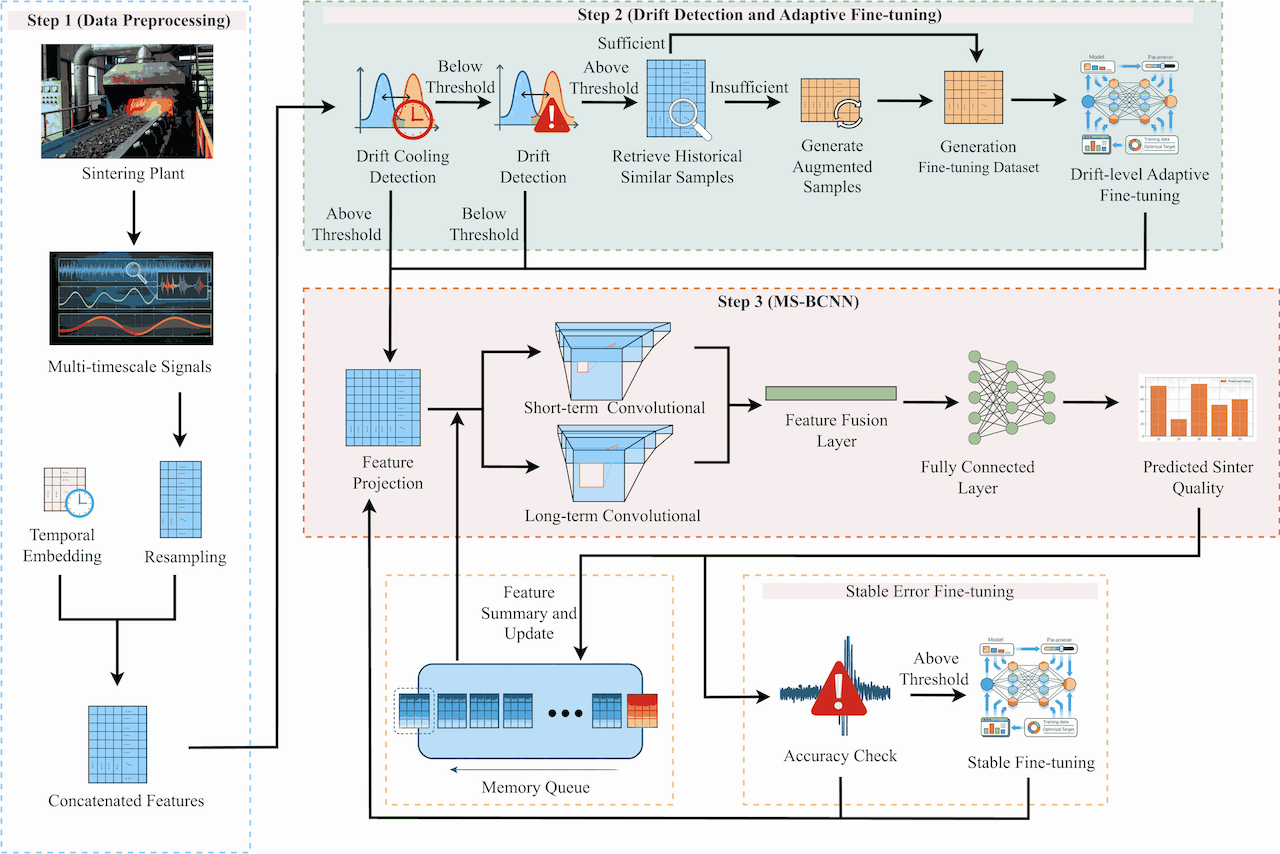}
\caption{Overall architecture of the DA-MSDL framework. The pipeline integrates Data Preprocessing (Step 1), Drift Detection and Adaptive Fine-tuning (Step 2), and MS-BCNN Feature Extraction (Step 3), supported by the Dynamic Memory Queue and Stable Error Fine-tuning modules.}
\label{fig:framework}
\end{figure*}

As shown in Fig.~\ref{fig:framework}, DA-MSDL follows a proactive detect--adapt--predict pipeline: Step~1 preprocesses the stream and constructs a sliding detection window, Step~2 performs drift-aware triggering via unsupervised discrepancy measures and activates hierarchical fine-tuning, and Step~3 extracts multi-scale features with MS-BCNN for forecasting.
The Dynamic Memory Queue supplies cross-temporal context to cope with fast distribution evolution, while Stable Error Fine-tuning calibrates residual bias once delayed labels return, jointly balancing adaptation responsiveness and forgetting mitigation under delayed supervision.

\subsection{Dynamic Memory Queue}
\label{sec:memory_queue}

To provide the model with cross-temporal context during online adaptation, this section introduces a dynamic memory queue mechanism.
By maintaining a sliding queue of historical feature summaries, this module explicitly represents recent operating-condition evolution and supports cross-temporal context modeling for online forecasting.

\subsubsection{Memory Construction and Aggregation}

DA-MSDL constructs a Dynamic Memory Queue based on high-level backbone features to preserve compact summaries of recent operating conditions.
To align the raw input dimension $F$ with the latent dimension of the backbone, the process feature vector $\mathbf{x}_t \in \mathbb{R}^{F}$ is first projected into a $C$-dimensional space:
\begin{equation}
    \mathbf{z}_t = \mathbf{W}_p \mathbf{x}_t + \mathbf{b}_p,
    \quad \mathbf{W}_p \in \mathbb{R}^{C\times F},\ \mathbf{b}_p \in \mathbb{R}^{C},
\end{equation}
where $\mathbf{z}_t \in \mathbb{R}^{C}$ denotes the projected feature, and $\mathbf{W}_p$ and $\mathbf{b}_p$ are the learnable weight matrix and bias vector of the projection layer, respectively.
Note that $\mathbf{x}_t$ and $\mathbf{z}_t$ lie in the raw feature space and the representation space, respectively: $\mathbf{x}_t$ is the raw process-variable vector, while $\mathbf{z}_t$ serves as the input representation to the backbone.
The backbone then encodes the projected sequence $\{\mathbf{z}_{t-L+1},\ldots,\mathbf{z}_t\}$ to extract high-level features, based on which a compact memory item is constructed and stored in the queue.
At time step $t$, the model maintains a memory queue $\mathcal{M}_t$ with maximum length $T_m$, containing compact memory items from previous time steps:
\begin{equation}
\begin{aligned}
\mathcal{M}_t &= [\mathbf{m}_{t-|\mathcal{M}_t|}, \ldots, \mathbf{m}_{t-2}, \mathbf{m}_{t-1}], \\
&\quad |\mathcal{M}_t| \le T_m,\quad \mathbf{m}_i \in \mathbb{R}^{C}.
\end{aligned}
\label{eq:memory_queue_def}
\end{equation}
where $\mathbf{m}_i \in \mathbb{R}^{C}$ denotes a compact memory item derived from backbone features at time step $i$, and the queue is ordered from the oldest item to the most recent item.
To maintain a fixed memory budget, the queue is updated in a first-in-first-out (FIFO) manner, where the oldest item is discarded whenever a new item is inserted after the queue reaches length $T_m$.

To obtain a stable contextual representation, the model aggregates the most recent memory items by average pooling. The effective aggregation length is defined as:
\begin{equation}
    R_t = \min(R, |\mathcal{M}_t|), \qquad 1 \le R \le T_m,
\end{equation}
and the aggregated memory $\bar{\mathbf{m}}_t$ is defined as:
\begin{equation}
    \bar{\mathbf{m}}_t =
    \begin{cases}
        \dfrac{1}{R_t}\sum_{i=1}^{R_t}\mathbf{m}_{t-i}, & R_t > 0, \\[6pt]
        \mathbf{0}, & R_t = 0.
    \end{cases}
\end{equation}
Here, $\mathbf{z}_t$ represents the instantaneous operating state of the current sample, while $\bar{\mathbf{m}}_t$ summarizes the recent historical context.

\subsubsection{Gated Feature Fusion and Update Loop}

To adaptively balance instantaneous information and historical context, DA-MSDL employs a linear gated fusion mechanism.
The gate vector $\mathbf{g}_t$ is computed from the current projected feature $\mathbf{z}_t$ and the aggregated memory $\bar{\mathbf{m}}_t$:
\begin{equation}
    \mathbf{g}_t = \sigma\!\left(\mathbf{W}_g[\mathbf{z}_t;\bar{\mathbf{m}}_t] + \mathbf{b}_g\right),
    \quad \mathbf{g}_t \in (0,1)^C,
\end{equation}
where $[\cdot;\cdot]$ denotes vector concatenation, $\sigma(\cdot)$ is the sigmoid function, and $\mathbf{W}_g,\mathbf{b}_g$ are learnable parameters. 
The memory-enhanced feature $\tilde{\mathbf{z}}_t$ is then constructed as:
\begin{equation}
    \tilde{\mathbf{z}}_t = \mathbf{g}_t \odot \mathbf{z}_t + (\mathbf{1}-\mathbf{g}_t)\odot \bar{\mathbf{m}}_t,
\end{equation}
where $\odot$ denotes the Hadamard product.

For forecasting at time $t$, the memory-enhanced features over the latest $L$ steps are stacked into an input window:
\begin{equation}
    \tilde{\mathbf{Z}}_{t-L+1:t}
    =
    \left[\tilde{\mathbf{z}}_{t-L+1},\tilde{\mathbf{z}}_{t-L+2},\ldots,\tilde{\mathbf{z}}_t\right]
    \in \mathbb{R}^{C\times L}.
\end{equation}
In each input window $\tilde{\mathbf{Z}}_{t-L+1:t}$, each column corresponds to one time step; thus, the first dimension indexes feature channels and the second dimension indexes temporal order.
The current stacked window  is then processed by the MS-BCNN backbone to extract multi-scale spatiotemporal representations:
\begin{equation}
\begin{aligned}
\mathbf{h}_t
&= \mathcal{F}_{\mathrm{backbone}}\!\left(\tilde{\mathbf{Z}}_{t-L+1:t};\theta_t\right) \\
&= \mathrm{Concat}\!\left(
\mathcal{F}_{\mathrm{short}}\!\left(\tilde{\mathbf{Z}}_{t-L+1:t};\theta_t\right),
\mathcal{F}_{\mathrm{long}}\!\left(\tilde{\mathbf{Z}}_{t-L+1:t};\theta_t\right)
\right).
\end{aligned}
\end{equation}
where $\mathcal{F}_{\mathrm{short}}$ and $\mathcal{F}_{\mathrm{long}}$ denote the short- and long-kernel convolution branches, respectively, and $\mathrm{Concat}(\cdot)$ denotes channel-wise concatenation.

After prediction, the memory state is updated. 
To compress the high-level representation into a compact memory item, the model applies global average pooling (GAP), followed by a memory-space projection with ReLU activation:
\begin{equation}
    \mathbf{m}_t = \mathrm{ReLU}\!\left(\mathbf{W}_m\,\mathrm{GAP}(\mathbf{h}_t) + \mathbf{b}_m\right),
    \quad \mathbf{m}_t \in \mathbb{R}^{C},
\end{equation}
where $\mathbf{W}_m$ and $\mathbf{b}_m$ are the learnable parameters of the memory projection layer, which map the pooled feature to the $C$-dimensional memory space.

The newly generated memory item $\mathbf{m}_t$ is written into the Dynamic Memory Queue, and the next-step memory state $\mathcal{M}_{t+1}$ is obtained by the FIFO update rule defined in Eq.~(\ref{eq:memory_queue_def}).
This mechanism maintains continuous cross-temporal context and provides stable representational support for subsequent drift detection and adaptive fine-tuning.

\subsection{Unsupervised Drift Awareness and Detection}
\label{sec:drift_aware}

This subsection describes the workflow of the unsupervised drift detection mechanism.
The decision process consists of two stages. First, drift is detected in the feature space by comparing the current observations with a reference context.
Second, local distributional characteristics are characterized using a sliding-window scheme, and the resulting drift severity signal is passed to the subsequent adaptive fine-tuning module.

\subsubsection{Unsupervised Drift Indicator via MMD}

To quantify the discrepancy between the current operating condition and the reference condition associated with the current model state, DA-MSDL adopts Maximum Mean Discrepancy (MMD)~\cite{Gretton2012Kernel} as the unsupervised drift indicator.
MMD quantifies distribution shift by measuring the distance between the mean embeddings of two sample sets in a reproducing kernel Hilbert space (RKHS). Its squared empirical form is defined as:
\begin{align}
\mathrm{MMD}^2(\mathbf{Z}^{\mathrm{ref}}_t, \mathbf{Z}^{\mathrm{cur}}_t)
&=
\frac{1}{L_w^{2}}
\sum_{i=1}^{L_w}\sum_{j=1}^{L_w}
k(\mathbf{u}^{\mathrm{ref}}_{t,i},\mathbf{u}^{\mathrm{ref}}_{t,j}) \notag\\
&\quad+ \frac{1}{L_w^{2}}
\sum_{i=1}^{L_w}\sum_{j=1}^{L_w}
k(\mathbf{u}^{\mathrm{cur}}_{t,i},\mathbf{u}^{\mathrm{cur}}_{t,j}) \notag\\
&\quad- \frac{2}{L_w^{2}}
\sum_{i=1}^{L_w}\sum_{j=1}^{L_w}
k(\mathbf{u}^{\mathrm{ref}}_{t,i},\mathbf{u}^{\mathrm{cur}}_{t,j}),
\label{eq:mmd}
\end{align}
where $k(\cdot,\cdot)$ denotes the Gaussian kernel:
\begin{equation}
k(\mathbf{a},\mathbf{b})=\exp\!\left(-\frac{\|\mathbf{a}-\mathbf{b}\|_2^2}{2\sigma^2}\right).
\end{equation}
The bandwidth $\sigma$ is set as $\sigma=\sqrt{C/2}$ according to the representation dimension used for MMD computation, where $C$ is the feature-representation dimension defined above.
Here, $\mathbf{Z}^{\mathrm{cur}}_t=\{\mathbf{u}^{\mathrm{cur}}_{t,i}\}_{i=1}^{L_w}$ denotes the current detection window and $\mathbf{Z}^{\mathrm{ref}}_t=\{\mathbf{u}^{\mathrm{ref}}_{t,i}\}_{i=1}^{L_w}$ denotes the reference window associated with the current model state, where $i$ indexes samples within each window.
This window-based construction ensures that drift estimation reflects local distributional variation rather than sample-wise fluctuations.
Specifically, $\mathbf{Z}^{\mathrm{cur}}_t$ is constructed by a sliding window over newly arrived unlabeled samples, while $\mathbf{Z}^{\mathrm{ref}}_t$ characterizes the distribution to which the model was most recently adapted during fine-tuning.
Accordingly, $\mathbf{Z}^{\mathrm{ref}}_t$ is updated only when model adaptation is triggered, rather than at every time step.
As a result, not every current detection window is promoted to the next reference window.
Only after a drift-triggered fine-tuning step is completed is the current detection window $\mathbf{Z}^{\mathrm{cur}}_t$ recorded as the reference window for subsequent drift estimation.

\subsubsection{Adaptive Sliding Window and Drift Categorization}

To characterize local distributional variation for MMD-based drift estimation, the current detection window, denoted by $\mathbf{Z}^{\mathrm{cur}}_t$, is constructed from the current sample and its preceding samples.
As the comparative baseline, the reference window, denoted by $\mathbf{Z}^{\mathrm{ref}}_t$, is defined as the detection window associated with the most recent model adaptation.
This design ensures that drift is consistently measured between the current sample distribution and the distribution to which the model is currently aligned, thereby maintaining a one-to-one correspondence between drift evaluation and the model state.

The drift severity level $d_t \in \{0,1,2,3\}$ is obtained by mapping the MMD score
$V_t=\mathrm{MMD}^2(\mathbf{Z}^{\mathrm{ref}}_t,\mathbf{Z}^{\mathrm{cur}}_t)$
to a set of predefined thresholds:
\begin{equation}
d_t =
\begin{cases}
0, & V_t < \lambda_{\mathrm{mild}},\\
1, & \lambda_{\mathrm{mild}} \le V_t < \lambda_{\mathrm{mod}},\\
2, & \lambda_{\mathrm{mod}} \le V_t < \lambda_{\mathrm{sev}},\\
3, & V_t \ge \lambda_{\mathrm{sev}}.
\end{cases}
\label{eq:drift_categorization}
\end{equation}
where $\lambda_{\mathrm{mild}}$, $\lambda_{\mathrm{mod}}$, and $\lambda_{\mathrm{sev}}$ denote the sensitivity thresholds for mild, moderate, and severe drift levels, respectively.
Here, $d_t=0$ indicates that no drift-triggered adaptation is performed at time $t$, while $d_t \in \{1,2,3\}$ corresponds to mild, moderate, and severe drift-triggered adaptation, respectively.

To avoid frequent triggers caused by noise, MMD is evaluated only when sufficient samples are available to form a valid detection window and the detection cooldown condition is satisfied, i.e., $t - t_{\mathrm{last}} \ge T_{\mathrm{cool}}$, where $t_{\mathrm{last}}$ denotes the time step of the most recent adaptation and $T_{\mathrm{cool}}$ is the minimum adaptation interval.
Because the online adaptation process may be unstable during the initial stage, DA-MSDL imposes an additional drift-severity constraint during early drift-triggered adaptation events.
Specifically, let $C_t$ denote the cumulative number of drift events that trigger model adaptation. When $C_t \le N_{\mathrm{init}}$, the effective drift level is capped at 1:
\begin{equation}
d_t = \min(d_t, 1).
\label{eq:initial_cap}
\end{equation}
This constraint suppresses overly aggressive updates caused by local transients in the early online evolution stage, thereby promoting a smoother adaptation process.

\subsection{Trend-Aware Hierarchical Fine-tuning}
\label{sec:fine_tuning}

This subsection details the trend-aware hierarchical fine-tuning mechanism, which adjusts the model update path according to the detected drift severity level $d_t$.
The fine-tuning mechanism consists of three components: first, a joint loss function that captures prediction error, directional trend, and volatility is used to guide adaptation.
Second, hierarchical parameter updates are performed together with multi-scale sample augmentation to accelerate alignment with the shifted distribution.
Finally, a stable error fine-tuning mechanism is introduced to correct long-term accumulated bias when no explicit drift is detected.

\subsubsection{Trend-Aware Joint Loss Function}

To reduce oversensitivity to short-term noise, we introduce a trend-aware joint loss that combines prediction error, trend consistency, difference-magnitude consistency, and volatility consistency, with drift-level-dependent weighting to control the fine-tuning intensity.
The trend-aware joint loss at time $t$ is defined as
\begin{equation}
\mathcal{L}_t^{(d_t)} =
\mathcal{L}_{\mathrm{err}}
+ w_{\mathrm{tr}}^{(d_t)} \mathcal{L}_{\mathrm{trend}}
+ w_{\mathrm{df}}^{(d_t)} \mathcal{L}_{\mathrm{diff}}
+ w_{\mathrm{vl}}^{(d_t)} \mathcal{L}_{\mathrm{vol}},
\label{eq:joint_loss}
\end{equation}
where $\mathcal{L}_{\mathrm{err}}$ denotes the base prediction error term, and $w_{\mathrm{tr}}^{(d_t)}$, $w_{\mathrm{df}}^{(d_t)}$, and $w_{\mathrm{vl}}^{(d_t)}$ are drift-level-dependent weights for the trend, difference, and volatility consistency terms, respectively.
The auxiliary terms are computed over a short supervised sequence of length $H$; this sequence is formed by temporally ordered predictions and ground-truth values in the output space for joint-loss modeling and is distinct from the feature-memory items in the Dynamic Memory Queue.
Specifically, these auxiliary terms are averaged across the $K$ output dimensions and are defined as:
\begin{equation}
\mathcal{L}_{\mathrm{trend}}
=
\frac{1}{K(H-1)}
\sum_{k=1}^{K}\sum_{\tau=t-H+2}^{t}
\left(\Delta \hat{y}_{\tau}^{(k)} - \Delta y_{\tau}^{(k)}\right)^2
\label{eq:l_trend}
\end{equation}
\begin{equation}
\mathcal{L}_{\mathrm{diff}}
=
\frac{1}{K(H-2)}
\sum_{k=1}^{K}\sum_{\tau=t-H+3}^{t}
\left(
\Delta^2 \hat{y}_{\tau}^{(k)} - \Delta^2 y_{\tau}^{(k)}
\right)^2
\label{eq:l_diff}
\end{equation}
\begin{equation}
\mathcal{L}_{\mathrm{vol}}
=
\frac{1}{K}
\sum_{k=1}^{K}
\left(
\mathrm{Var}\!\left(\hat{\mathbf{y}}_{t-H+1:t}^{(k)}\right)
-
\mathrm{Var}\!\left(\mathbf{y}_{t-H+1:t}^{(k)}\right)
\right)^2
\end{equation}
where $\Delta y_{\tau}^{(k)} = y_{\tau}^{(k)} - y_{\tau-1}^{(k)}$ and $\Delta \hat{y}_{\tau}^{(k)} = \hat{y}_{\tau}^{(k)} - \hat{y}_{\tau-1}^{(k)}$ denote the first-order temporal differences, and $\Delta^2 y_{\tau}^{(k)} = \Delta y_{\tau}^{(k)} - \Delta y_{\tau-1}^{(k)}$ and $\Delta^2 \hat{y}_{\tau}^{(k)} = \Delta \hat{y}_{\tau}^{(k)} - \Delta \hat{y}_{\tau-1}^{(k)}$ denote the second-order temporal differences; the former and latter are used for trend-consistency and local fluctuation-consistency modeling, respectively. Here, $\hat{\mathbf{y}}_{t-H+1:t}^{(k)}$ and $\mathbf{y}_{t-H+1:t}^{(k)}$ denote the predicted and ground-truth subsequences of the $k$-th output over the same window.
As the detected drift level $d_t$ at time $t$ increases, the drift-level-dependent weights $w_{\mathrm{tr}}^{(d_t)}$, $w_{\mathrm{df}}^{(d_t)}$, and $w_{\mathrm{vl}}^{(d_t)}$ in \eqref{eq:joint_loss} are progressively adjusted to place greater emphasis on trend- and fluctuation-related consistency constraints, thereby enabling more aggressive local temporal realignment under stronger drift.

\subsubsection{Hierarchical Parameter Update and Sample Augmentation}

To enable controlled model updating, DA-MSDL adaptively modulates the fine-tuning intensity according to the detected drift severity $d_t \in \{1,2,3\}$.
The online parameter optimization follows a drift-level-dependent update rule over the active trainable parameters:
\begin{equation}
\theta_{t+1}^{\mathrm{act}} = \theta_t^{\mathrm{act}} - \eta_{d_t}\nabla_{\theta_t^{\mathrm{act}}}\mathcal{L}_t^{(d_t)},
\label{eq:parameter_update}
\end{equation}
where $\theta_t^{\mathrm{act}}$ denotes the subset of parameters activated for updating at drift level $d_t$, while all frozen parameters remain unchanged.
The step size $\eta_{d_t}$ is determined by the drift level, and $\nabla_{\theta_t^{\mathrm{act}}}\mathcal{L}_t^{(d_t)}$ is the gradient of the joint loss in Eq.~(\ref{eq:joint_loss}) with respect to the active parameters.

For different drift levels, the hierarchical adaptation strategy is instantiated through level-specific configurations of trainable layers, optimization intensity, and adaptation data composition.
The resulting drift-severity-aware hierarchical fine-tuning mechanism can be viewed as a form of cross-condition transfer learning, in which the previously adapted model state is reused as initialization and progressively adapted to the current operating condition.
For mild drift ($d_t=1$), the system performs low-intensity local correction. 
The backbone remains frozen, and only the prediction head is updated. 
Fine-tuning uses the smallest learning rate $\eta_{1}$ and iteration budget $T_{1}$, together with the lowest drift-aware weight set $\mathbf{w}^{(1)}=[w_{\mathrm{tr}}^{(1)},w_{\mathrm{df}}^{(1)},w_{\mathrm{vl}}^{(1)}]$, to suppress short-term noise.
For moderate drift ($d_t=2$), the unfreezing boundary is expanded to the high-level convolutional blocks and the fusion layer of the backbone. 
Accordingly, the learning rate $\eta_{2}$, iteration budget $T_{2}$, and drift-aware weight set $\mathbf{w}^{(2)}$ are moderately increased relative to Level 1, enabling the model to capture structural changes in local operating conditions.
For severe drift ($d_t=3$), the system triggers full-parameter adaptation by unfreezing all learnable weights. 
To handle substantial distributional shifts, the strongest hyperparameter configuration $\{\eta_{3}, T_{3}, \mathbf{w}^{(3)}\}$ is applied, promoting rapid alignment with the latest target distribution through more aggressive optimization and stronger trend-aware guidance.

The supervised adaptation set is constructed to provide sufficient training support for online fine-tuning when immediately usable labeled samples are limited, with a minimum size of $N_{\mathrm{ft}}$.
Due to label verification latency, newly arrived samples in the current detection window typically do not have labels yet and therefore cannot be directly used for supervised fine-tuning.
In contrast, samples in the same window whose labels have already become available remain eligible for inclusion.
The adaptation set is constructed by filling samples in a priority order: (i) labeled samples available in the current detection window, (ii) historical labeled samples selected by distributional similarity to the current detection window, (iii) their secondary resampled variants, and (iv) feature-perturbation augmented samples.
Samples are appended sequentially following the above order until the set size reaches $N_{\mathrm{ft}}$.

To preserve historical labeled samples for supervised adaptation, the system maintains a historical replay buffer with maximum capacity $N_{\mathrm{buf}}$, which is distinct from the Dynamic Memory Queue used for feature-context modeling.
Historical similar samples are retrieved by traversing the replay buffer to construct candidate labeled windows, where the candidate windows use the same window length as the current detection window.
For each candidate labeled window $\mathbf{Z}^{(j)}_{\mathrm{buf}}$, the similarity score is computed as $\mathrm{MMD}^2(\mathbf{Z}^{(j)}_{\mathrm{buf}}, \mathbf{Z}^{\mathrm{cur}}_t)$ in the same feature space, and only candidates with scores below $\tau_h$ are retained.
Secondary resampling is applied in accordance with the multi-scale structure of MS-BCNN, where a resampling operator (e.g., linear interpolation, pooling, or anti-aliasing convolution) is randomly selected to generate additional scale variants. 
Furthermore, feature-perturbation augmented samples are generated from the selected supervised adaptation samples by injecting Gaussian perturbations into their multi-scale feature representations:
\begin{equation}
\mathbf{z}_{\mathrm{gen}} = \mathbf{z} + \epsilon\boldsymbol{\xi}, \quad \boldsymbol{\xi} \sim \mathcal{N}\!\left(\mathbf{0}, \Sigma\right),
\label{eq:gaussian_augmentation}
\end{equation}
where $\mathbf{z}$ and $\mathbf{z}_{\mathrm{gen}}$ denote the original and augmented feature vectors of a supervised adaptation sample, respectively, and $\boldsymbol{\xi}$ is a Gaussian noise vector with the same dimension as $\mathbf{z}$. Here, $\epsilon$ is the drift-level-dependent perturbation magnitude, and $\Sigma$ is the diagonal covariance matrix estimated from the currently available supervised adaptation samples in the same feature space.
This hybrid sample construction yields an adaptation set that combines current-context supervision, distribution-matched historical samples, and multi-scale feature augmentations.

\subsubsection{Stable Error Fine-tuning}

In addition to drift-driven fine-tuning, DA-MSDL introduces an error-driven stable fine-tuning branch for the no-drift state ($d_t=0$) to correct gradually accumulated predictive bias.
A stabilized error sequence is tracked using an exponential moving average (EMA):
\begin{equation}
\hat{e}_t = (1-\lambda_e)\hat{e}_{t-1} + \lambda_e e_t, \quad \lambda_e \in (0,1],
\label{eq:ema_error}
\end{equation}
where $e_t$ denotes the window-level prediction error computed on the current detection window $\mathbf{Z}^{\mathrm{cur}}_t$ when the corresponding delayed labels become available, and $\lambda_e$ is the smoothing coefficient.
The stable-error branch is defined on the same sliding detection window $\mathbf{Z}^{\mathrm{cur}}_t$ used in drift detection.
Specifically, $e_t$ is defined as the window-level MAE on $\mathbf{Z}^{\mathrm{cur}}_t$:
\begin{equation}
e_t=
\frac{1}{L_w K}\sum_{i=1}^{L_w}\left\|\hat{\mathbf{y}}_{t,i}-\mathbf{y}_{t,i}\right\|_1,
\label{eq:window_error}
\end{equation}
where $L_w$ is the detection-window length and $K$ is the number of prediction targets.
Here, $\hat{\mathbf{y}}_{t,i}$ and $\mathbf{y}_{t,i}$ denote the prediction and ground truth of the $i$-th sample in the window $\mathbf{Z}^{\mathrm{cur}}_t$, respectively.
When the number of available samples in the current detection window is insufficient, the stable-error branch skips error evaluation at time $t$ and keeps the previous EMA state unchanged.

To reduce computational overhead and suppress transient noise, stable fine-tuning is triggered only when the system is in the no-drift state ($d_t=0$) and the smoothed error $\hat{e}_t$ exceeds a predefined threshold $\tau_e$ for $K_e$ consecutive steps.
Once triggered, stable fine-tuning uses a dedicated learning rate $\eta_e$ and a minimum of $T_e$ optimization iterations to update only the prediction head, while keeping the MS-BCNN backbone frozen.

\section{Experiments}
\label{Section4}

This section establishes the experimental foundation by analyzing the nonstationary characteristics of the selected industrial datasets, followed by a systematic validation under defined parameter configurations. 
Integrating results from both comparative and ablation studies, this section comprehensively demonstrates the predictive superiority of DA-MSDL and the necessity of its core adaptive modules.

\subsection{Dataset Description and Analysis}

\subsubsection{Data Description}

The experimental dataset was collected from the process monitoring system and laboratory assay system of a sintering plant at a large iron and steel enterprise in China, covering the operating conditions of a belt sintering machine in real production.
For online quality modeling and dynamic prediction, multiple sensors and metering devices are deployed throughout the production line.
The monitored variables include process features such as mixed-material moisture, temperature, particle size distribution, and ignition temperature, as well as quality indicators obtained from end-point laboratory assays, including sinter chemical compositions (TFe, FeO, SiO$_2$, and CaO) and basicity.
The sampling intervals of these variables span a wide range, from seconds to months. Accordingly, they are grouped into high-, medium-, and low-frequency categories based on their temporal resolutions.
Given that a complete sintering cycle lasts approximately 2~h, we segmented and time-aligned approximately seven months of continuous production records in a batch-wise manner, such that each sample corresponds to one full cycle, yielding 2,283 valid samples.
The feature categories and representative variables are summarized in Table \ref{tab:feature_categories}.

\begin{table}[!t]
\caption{Feature categories and representative variables in the sintering dataset.}
  \begin{tabularx}{\linewidth}{p{2.2cm} >{\raggedright\arraybackslash}X >{\raggedright\arraybackslash}X}
    \toprule
    Category & Representative Features & Physical Meaning \\
    \midrule

    Raw Materials \& Blending &
    Mixed-material moisture, temperature, and particle size distribution; \newline
    fuel particle size; \newline
    ore blend proportions 
    &
    Describe the raw-material composition and blending conditions before sintering \\
    \midrule

    Process \& Thermal Parameters &
    Ignition temperature; \newline
    exhaust gas temperature; \newline
    burn-through~point (BTP); \newline
    main-line and furnace negative pressures; \newline
    coal-gas flow rate and pressure; air flow rate and pressure
    &
    Reflect combustion intensity, heat-transfer behavior, and airflow dynamics \\
    \midrule

    Equipment \& Environment &
    Sintering machine speed; pallet speed; \newline
    bed height; finished-product weight; \newline
    production output
    &
    Indicate equipment operating status and system-level production conditions \\
    \midrule

    Output Targets &
    TFe, FeO, SiO$_2$, CaO, basicity &
    Key chemical quality indicators obtained from laboratory assays \\
    \bottomrule
  \end{tabularx}
  \label{tab:feature_categories}
\end{table}

\subsubsection{Data Preprocessing}

Before model training, the raw production data were cleaned and examined to ensure data quality and consistency.
Specifically, samples collected during equipment start-up and shutdown phases were excluded because these periods exhibit nonstationary transient behaviors that are not representative of stable production.
For sporadic missing values caused by temporary sensor disconnections, linear interpolation was applied to recover local continuity in the time series.
Subsequently, all process features were time-aligned to a unified 2-h production cycle and normalized using min-max scaling to mitigate scale differences across variables, while the quality targets were kept in their original scales for model training and evaluation.

\subsubsection{Correlation and Coupling Analysis}

\begin{figure}[!t]
\centering
\includegraphics[width=\linewidth]{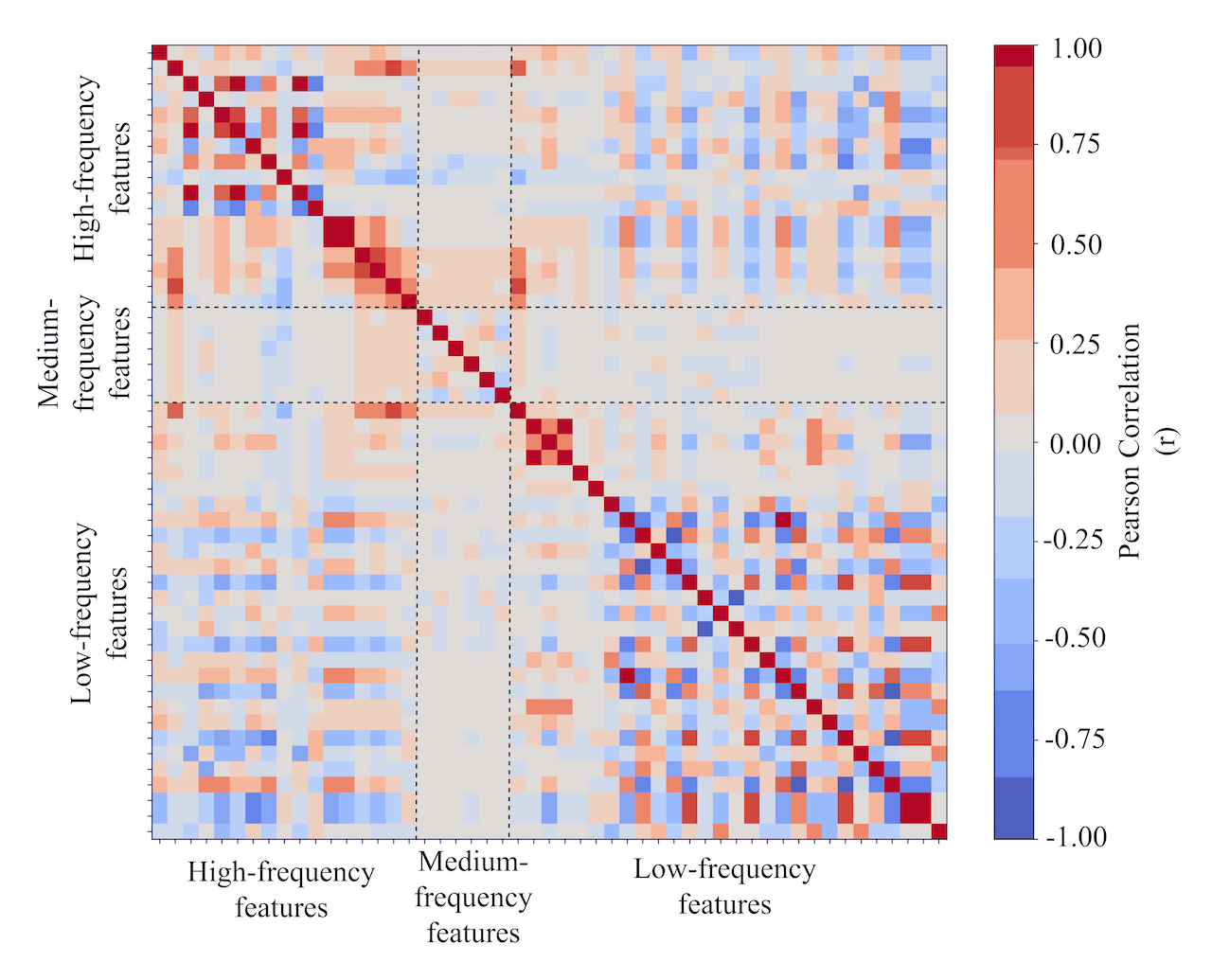}
\caption{Feature-feature correlation heatmap indicating spatial locality and frequency-based coupling.}
\label{fig:feature_corr}
\end{figure}

The Pearson correlation coefficient matrix was computed to characterize the coupling structure among process features.
Fig. \ref{fig:feature_corr} shows the correlation heatmap of process features after preprocessing, where the dotted partitions indicate feature groups with different temporal resolutions.
High-frequency features exhibit strong positive correlations, which reflect the physical couplings among temperature, airflow, and pressure variables.
Moreover, the localized block patterns in cross-group feature correlations suggest local dependency structures across feature groups, which motivates the use of multi-channel convolutions to capture cross-feature local interactions.

\begin{figure}[!t]
\centering
\includegraphics[width=\linewidth]{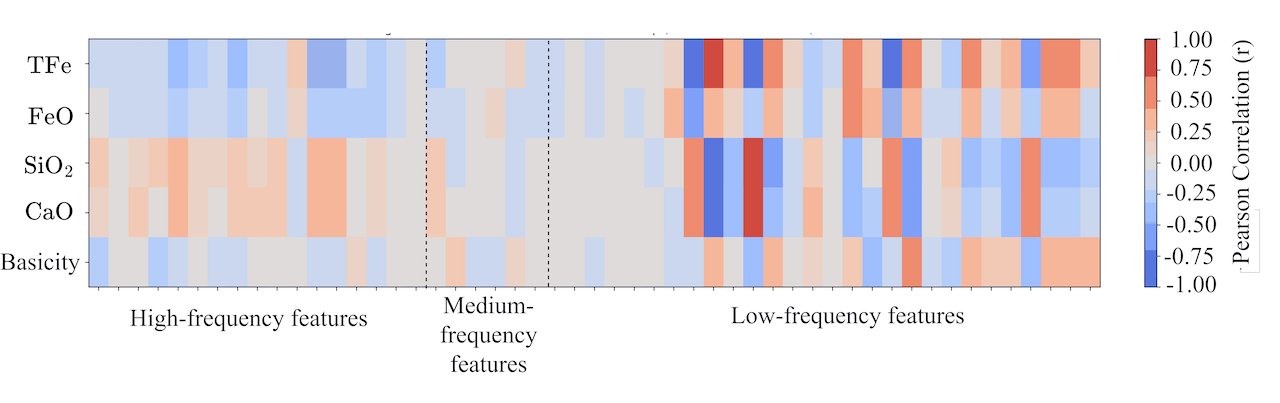}
\caption{Feature-target correlation heatmap illustrating complex dependencies.}
\label{fig:label_corr}
\end{figure}

Furthermore, Fig. \ref{fig:label_corr} presents the correlation heatmap between process features and quality targets.
Raw-material blending features exhibit consistently positive correlations with CaO content and basicity, whereas some thermal-process features show negative correlations with FeO content.
These correlation patterns indicate complex dependencies jointly shaped by multi-scale dynamics and multiple process factors in the sintering process.

\subsubsection{Non-stationarity and Distribution Drift}

\begin{figure}[!t]
\centering
\includegraphics[width=\linewidth]{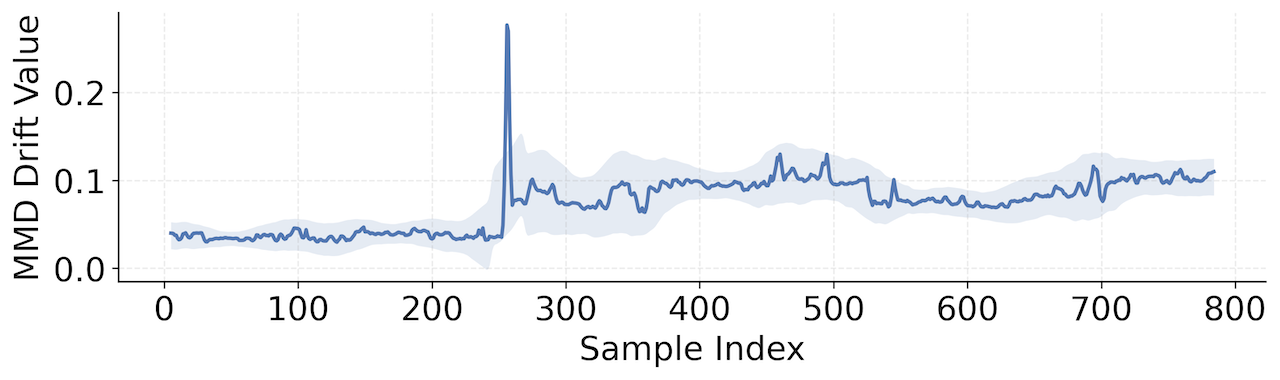}
\caption{MMD-based drift trend during the online detection stage.}
\label{fig:mmd_drift}
\end{figure}

To quantify distribution drift over time, the dataset is divided into an offline training phase and an online detection phase.
For each detection step, the MMD with respect to the offline training set is computed. As shown in Fig. \ref{fig:mmd_drift}, the MMD curve exhibits an overall increasing trend over time, together with several abrupt spikes, indicating persistent distribution drift and nonstationarity in the process features.
These pronounced distribution shifts suggest that static models may experience performance degradation during long-term deployment, thereby providing empirical motivation for the proposed dynamic adaptive framework, DA-MSDL.

\subsection{Experimental Setup}

\subsubsection{Parameter Settings}

The proposed DA-MSDL framework is implemented on a local computing platform equipped with an Apple M4 Pro chip (16GB RAM).
This environment supports rapid iteration and real-time debugging, which is suitable for evaluating the computational behavior of online adaptive algorithms in industrial settings.
All experiments are implemented in PyTorch, and model parameters are optimized using AdamW.

Regarding the MS-BCNN backbone architecture and the offline training configuration, we follow the settings reported previously~\cite{mypaper}. 
Specifically, the input sequence length is set to $L=12$, and the number of output targets is $K=5$.
The initial offline modeling phase uses the first 1,500 samples, with a batch size of 32.
For the Dynamic Memory Queue module, the queue capacity is set to $T_m=4$, and all memory items are used for aggregation, i.e., the effective aggregation length is set to $R=T_m$.

For drift detection, the sliding detection window length is set to $L_w=5$, and the detection cooldown interval is set to $T_{\mathrm{cool}}=3$.
The drift severity thresholds in Eq.~\eqref{eq:drift_categorization} are set to $\lambda_{\mathrm{mild}}=0.05$, $\lambda_{\mathrm{mod}}=0.12$, and $\lambda_{\mathrm{sev}}=0.2$.
The initial severity-cap count in Eq.~\eqref{eq:initial_cap} is set to $N_{\mathrm{init}}=3$.

For drift-triggered hierarchical fine-tuning, the supervised short-sequence length in the trend-aware joint loss of Eq.~\eqref{eq:joint_loss} is set to $H=8$.
In supervised adaptation-set construction, the minimum adaptation-set size is set to $N_{\mathrm{ft}}=300$ and the replay-buffer capacity is set to $N_{\mathrm{buf}}=800$; for similarity-based historical sample retrieval, the MMD similarity threshold is set to $\tau_h=0.05$.
For the feature-perturbation augmentation in Eq.~\eqref{eq:gaussian_augmentation}, the drift-level-dependent perturbation magnitude is implemented as a fixed setting in our experiments, with $\epsilon=0.01$.

For mild drift ($d_t=1$), the learning-rate scaling factor is $\eta_{1}=0.10$, the maximum fine-tuning epochs are $T_{1}=30$, the early-stopping patience is 5, and the validation split ratio is 0.15.
The lower-layer learning-rate multiplier is 0.5, the $L_2$-SP regularization coefficient is $5\times10^{-5}$, and the joint-loss weights are $\bigl(w_{\mathrm{tr}}^{(1)}, w_{\mathrm{df}}^{(1)}, w_{\mathrm{vl}}^{(1)}\bigr)=(0.3,\,0.2,\,0.05)$.
Only the prediction head is updated at this level, while the MS-BCNN backbone remains frozen.
For moderate drift ($d_t=2$), the learning-rate scaling factor is $\eta_{2}=0.15$, the maximum fine-tuning epochs are $T_{2}=40$, the early-stopping patience is 8, and the validation split ratio is 0.12.
The lower-layer learning-rate multiplier is 0.5, the $L_2$-SP regularization coefficient is $2\times10^{-6}$, and the joint-loss weights are $\bigl(w_{\mathrm{tr}}^{(2)}, w_{\mathrm{df}}^{(2)}, w_{\mathrm{vl}}^{(2)}\bigr)=(0.5,\,0.3,\,0.1)$.
Partial parameter unfreezing is applied at this level, where the high-level backbone blocks and fusion layers are trainable.
For severe drift ($d_t=3$), the learning-rate scaling factor is $\eta_{3}=0.25$, the maximum fine-tuning epochs are $T_{3}=50$, the early-stopping patience is 10, and the validation split ratio is 0.10.
The lower-layer learning-rate multiplier is 0.7, and the joint-loss weights are $\bigl(w_{\mathrm{tr}}^{(3)}, w_{\mathrm{df}}^{(3)}, w_{\mathrm{vl}}^{(3)}\bigr)=(0.7,\,0.4,\,0.2)$.
All learnable parameters are unfrozen at this level for full-parameter adaptation.

For stable error fine-tuning, an error-driven strategy is adopted on the same sliding detection window used in drift detection.
The primary trigger metric is MAE, the error threshold is set to $\tau_e=0.10$, the EMA smoothing coefficient is set to $\lambda_e=0.6$, and the consecutive trigger count $K_e$ is set to 2.
Once triggered, the stable-error branch uses a dedicated learning rate $\eta_e=0.1$ and a minimum optimization budget of $T_e=25$, while updating only the prediction head and keeping the MS-BCNN backbone frozen.

To protect industrial privacy, the raw plant data cannot be publicly released. For reproducibility, we provide an anonymized and normalized version of the dataset together with the implementation code and detailed preprocessing scripts.

\subsubsection{Baselines and Evaluation Metrics}

For model prediction performance evaluation, we compute Mean Squared Error (MSE)~\cite{Kong2025DLTSFSurvey}, Mean Absolute Error (MAE)~\cite{Kong2025DLTSFSurvey}, Mean Absolute Percentage Error (MAPE)~\cite{LimZohren2021TSFDeepLearningSurvey}, and the coefficient of determination ($R^2$)~\cite{LimZohren2021TSFDeepLearningSurvey} for each quality target.
These metrics are used to quantify prediction error and goodness of fit for each target. Specifically, smaller values of MSE, MAE, and MAPE indicate lower prediction errors, whereas an $R^2$ value closer to 1 indicates better fitting performance.
Considering the substantial differences in scales and numerical ranges across quality targets, directly averaging raw errors would bias the evaluation toward targets with larger magnitudes. Therefore, normalized mean squared error (NMSE)~\cite{Kong2025DLTSFSurvey} and normalized mean absolute error (NMAE)~\cite{LU2025ASurvey} are further adopted as scale-aware aggregate metrics, defined as follows:
\begin{equation}
\label{eq:normalized_metrics}
\begin{aligned}
\mathrm{NMSE} &= \frac{1}{K} \sum_{k=1}^{K} \frac{\mathrm{MSE}_k}{\sigma_k^2 + \epsilon_{reg}}, \\
\mathrm{NMAE} &= \frac{1}{K} \sum_{k=1}^{K} \frac{\mathrm{MAE}_k}{\sigma_k + \epsilon_{reg}}.
\end{aligned}
\end{equation}
where $\sigma_k$ denotes the standard deviation of the $k$-th target on the evaluation set, and $\epsilon_{reg}$ is a small positive constant introduced to prevent division by zero.
In addition, to facilitate overall comparison, the arithmetic means of MAPE and $R^2$ across the five quality targets ($K=5$) are reported as summary indicators of multi-target prediction performance.

For the multi-scale multi-target prediction task, five representative baseline models are selected for comparison.
All baseline methods are evaluated under the same dataset and experimental protocol, with configurations following the original references as closely as possible.
These baselines cover different modeling paradigms, including subspace identification, attention-based sequence modeling, and recurrent neural network architectures.

OB-ISSID~\cite{wang2025obissid} is a subspace-identification-based method for industrial processes, offering good interpretability and stable performance.
Ventingformer~\cite{dai2025ventingformer} combines convolutional feature extraction and attention mechanisms to model multi-scale temporal patterns and long-range dependencies in the sintering process.
Transformer~\cite{vaswani2017attention} uses self-attention to directly model global dependencies and serves as a representative architecture for sequence modeling.
LSTM~\cite{hochreiter1997lstm} mitigates vanishing gradients through gating mechanisms and can capture medium- and long-term dependencies in industrial time series.
GRU-PLS~\cite{yang2022grupls} integrates gated recurrent units with partial least squares regression, improving prediction robustness under noisy conditions while retaining nonlinear modeling capability.

\subsection{Experimental Results and Discussion}

\subsubsection{Shuffled Data Simulation}

To provide an upper-bound reference by largely removing distribution drift, we evaluate all methods on a randomly shuffled (approximately i.i.d.) version of the dataset, which assesses their fundamental ability to learn the static mapping from process features to multiple quality targets.

Table~\ref{tab:static_overall} summarizes the overall predictive performance on the shuffled dataset.
The results show that, under the approximate i.i.d. assumption, all baseline methods achieve reasonably good predictive accuracy.
Notably, the proposed DA-MSDL achieves the best performance in terms of the normalized metrics (NMSE and NMAE) and MAPE.
These results suggest that the multi-scale convolutional backbone can extract informative features under a stable distribution, providing a strong initialization for subsequent online adaptation.

Moreover, the competitive NMSE achieved by Transformer and Ventingformer indicates that the task is learnable under drift-free conditions.
This static benchmark serves as a reference for quantifying performance degradation due to distribution drift under real-world operating conditions.

\begin{table}[!t]
  \centering
  \caption{Overall predictive performance on the shuffled dataset (mean $\pm$ standard deviation).}
  \label{tab:static_overall}
  \begin{tabular}{lccc}
    \toprule
    \textbf{Model} & \textbf{NMSE} & \textbf{NMAE} & \textbf{MAPE (\%)} \\
    \midrule
    DA-MSDL
      & \cellcolor{gray!50}{$0.2423 \pm 0.1542$}
      & \cellcolor{gray!50}{$0.3052 \pm 0.1521$}
      & \cellcolor{gray!50}{$1.09 \pm 0.04$} \\
    OB-ISSID
      & $0.3044 \pm 0.2006$
      & $0.3809 \pm 0.1353$
      & $1.31 \pm 0.07$ \\
    Ventingformer
      & $0.2793 \pm 0.1989$
      & $0.3656 \pm 0.1364$
      & $1.25 \pm 0.07$ \\
    Transformer
      & $0.4564 \pm 0.1766$
      & $0.4829 \pm 0.1028$
      & $1.65 \pm 0.17$ \\
    LSTM
      & $0.3803 \pm 0.2204$
      & $0.4213 \pm 0.1419$
      & $1.44 \pm 0.08$ \\
    GRU-PLS
      & $0.4618 \pm 0.1707$
      & $0.5120 \pm 0.0821$
      & $1.74 \pm 0.08$ \\
    \bottomrule
  \end{tabular}
\end{table}

\subsubsection{Real-world Industrial Assessment}
\begin{table*}[t]
  \centering
 \caption{Predictive performance comparison on the five quality targets (mean $\pm$ standard deviation).}
  \label{tab:dynamic_full}
  \begin{tabular}{cccccc}
    \toprule
    \textbf{Model} & \textbf{Label} & \textbf{MSE} & \textbf{MAE} & \textbf{MAPE (\%)} & \boldmath{$R^2$} \\
    \midrule
    \multirow{6}{*}{DA-MSDL}
      & TFe      & \cellcolor{gray!50}0.060832 $\pm$ 0.006063 
                  & \cellcolor{gray!50}0.180206 $\pm$ 0.008479  
                  & \cellcolor{gray!50}0.320033 $\pm$ 0.015002 
                  & \cellcolor{gray!50}$-0.26131 \pm 0.125718$ \\
      & FeO      & \cellcolor{gray!50}0.118038 $\pm$ 0.006595 
                  & \cellcolor{gray!50}0.246699 $\pm$ 0.008104 
                  & \cellcolor{gray!50}2.682173 $\pm$ 0.089205 
                  & \cellcolor{gray!50}0.077400 $\pm$ 0.051546 \\
      & SiO$_2$  & \cellcolor{gray!50}0.006607 $\pm$ 0.001714 
                  & \cellcolor{gray!50}0.056760 $\pm$ 0.006537 
                  & \cellcolor{gray!50}1.028868 $\pm$ 0.117734 
                  & \cellcolor{gray!50}0.662256 $\pm$ 0.087604 \\
      & CaO      & \cellcolor{gray!50}0.036010 $\pm$ 0.003603 
                  & \cellcolor{gray!50}0.129945 $\pm$ 0.007947 
                  & \cellcolor{gray!50}1.092134 $\pm$ 0.067053 
                  & \cellcolor{gray!50}0.466945 $\pm$ 0.053337 \\
      & Basicity & \cellcolor{gray!50}0.000536 $\pm$ 0.000036 
                  & \cellcolor{gray!50}0.016871 $\pm$ 0.000561 
                  & \cellcolor{gray!50}0.784819 $\pm$ 0.026099 
                  & \cellcolor{gray!50}0.212011 $\pm$ 0.053578 \\
      & Overall Mean & \cellcolor{gray!50}{0.768453 $\pm$ 0.318736}
                  & \cellcolor{gray!50}{0.612549 $\pm$ 0.145467}
                  & \cellcolor{gray!50}{1.181605 $\pm$ 0.063019} 
                  & \cellcolor{gray!50}{0.231459 $\pm$ 0.074357} \\
    \midrule
    \multirow{5}{*}{OB-ISSID}
      & TFe      & 15.17148 $\pm$ 10.89632 & 3.240489 $\pm$ 1.500977 &  5.752892 $\pm$ 2.664363  & $-313.844 \pm 226.1248$ \\
      & FeO      &  7.427538 $\pm$ 11.39653 & 1.922083 $\pm$ 1.362099 & 21.09915 $\pm$ 14.89194 &  $-56.9798 \pm 88.96207$ \\
      & SiO$_2$  &  2.478063 $\pm$  2.177933 & 1.255659 $\pm$ 0.693490 & 22.82642 $\pm$ 12.58126 & $-125.770 \pm 111.4165$ \\
      & CaO      &  2.934882 $\pm$  1.721921 & 1.291617 $\pm$ 0.406435 & 10.96134 $\pm$  3.423734 &  $-42.5008 \pm 25.52234$ \\
      & Basicity &  0.037870 $\pm$  0.040804 & 0.154900 $\pm$ 0.084536 &  7.205134 $\pm$  3.930654 &  $-54.5410 \pm$ 59.84397 \\
      & Overall Mean & 119.6787 $\pm$ 101.7280
                  & 8.002796 $\pm$ 3.658696
                  & 13.56899 $\pm$ 7.498392 
                  & $-118.727 \pm 102.3739$ \\
    \midrule
    \multirow{5}{*}{Ventingformer}
      & TFe      & 0.618011 $\pm$ 0.430214 & 0.677437 $\pm$ 0.165248 & 1.201339 $\pm$ 0.954510 & $-9.74996 \pm 2.145630$  \\
      & FeO      & 0.344389 $\pm$ 0.161157 & 0.494081 $\pm$ 0.124562 & 5.503169 $\pm$ 1.158960 & $-1.68832 \pm 0.654242$  \\
      & SiO$_2$  & 0.049925 $\pm$ 0.004973 & 0.185943 $\pm$ 0.084562 & 3.437254 $\pm$ 1.425162 & $-1.55402 \pm 0.458556$  \\
      & CaO      & 0.079916 $\pm$ 0.014892 & 0.218651 $\pm$ 0.013254 & 1.851269 $\pm$ 0.242563 & $-0.18451 \pm 0.054621$  \\
      & Basicity & 0.002504 $\pm$ 0.000254 & 0.042743 $\pm$ 0.021445 & 1.976784 $\pm$ 0.581347 & $-2.67236 \pm 0.625842$  \\
      & Overall Mean & 4.584039 $\pm$ 4.191190
                  & 1.655030 $\pm$ 0.759926
                  & 2.793963 $\pm$ 0.872508 
                  & $-3.16983 \pm 0.787778$ \\
    \midrule
    \multirow{5}{*}{Transformer}
      & TFe      & 0.187186 $\pm$ 0.119301 & 0.348477 $\pm$ 0.125778 & 0.617939 $\pm$ 0.222995 & $-2.88456 \pm 2.475791$ \\
      & FeO      & 0.178953 $\pm$ 0.056598 & 0.317473 $\pm$ 0.061278 & 3.466190 $\pm$ 0.697501 & $-0.39691 \pm 0.441809$ \\
      & SiO$_2$  & 0.042787 $\pm$ 0.004404 & 0.170469 $\pm$ 0.009146 & 3.150201 $\pm$ 0.167836 & $-1.18885 \pm 0.225280$ \\
      & CaO      & 0.086698 $\pm$ 0.016337 & 0.228812 $\pm$ 0.021531 & 1.931392 $\pm$ 0.174227 & $-0.28503 \pm 0.242152$ \\
      & Basicity & 0.002679 $\pm$ 0.000277 & 0.044847 $\pm$ 0.002894 & 2.074490 $\pm$ 0.134311 & $-2.92924 \pm 0.405724$ \\
      & Overall Mean & 2.537357 $\pm$ 1.162249
                  & 1.258516 $\pm$ 0.346997
                  & 2.248042 $\pm$ 0.279374 
                  & $-1.53692 \pm 0.758151$ \\
    \midrule
    \multirow{5}{*}{LSTM}
      & TFe      & 1.543370 $\pm$ 1.162099  & 1.239216 $\pm$ 0.7562421 & 2.220838 $\pm$ 1.784521 & $-26.8781 \pm 2.145125$ \\
      & FeO      & 0.215657 $\pm$ 0.138634  & 0.363922 $\pm$ 0.026993  & 3.901176 $\pm$ 1.212547 &  $-0.68342 \pm 0.215479$ \\
      & SiO$_2$  & 0.049516 $\pm$ 0.062545  & 0.186542 $\pm$ 0.164988  & 3.451451 $\pm$ 1.567876 &  $-0.96584 \pm 0.312314$ \\
      & CaO      & 0.140146 $\pm$ 0.016552  & 0.348531 $\pm$ 0.112578  & 6.413463 $\pm$ 1.696815 &  $-6.16944 \pm 1.535647$ \\
      & Basicity & 0.358738 $\pm$ 0.148035  & 0.549547 $\pm$ 0.1935621 & 4.699463 $\pm$ 1.364844 &  $-4.31722 \pm 2.364879$ \\
      & Overall Mean & 114.7737 $\pm$ 206.4481
                  & 6.333591 $\pm$ 7.544510
                  & 4.227223 $\pm$ 1.525321 
                  & $-7.74633 \pm 1.314689$ \\
    \midrule
    \multirow{5}{*}{GRU-PLS}
      & TFe      & 0.943971 $\pm$ 0.046501 & 0.899228 $\pm$ 0.020768 & 1.596798 $\pm$ 0.036822 & $-18.5896 \pm 0.964998$ \\
      & FeO      & 0.387848 $\pm$ 0.036230 & 0.521086 $\pm$ 0.030214 & 5.807001 $\pm$ 0.337101 &  $-2.02756 \pm 0.282811$ \\
      & SiO$_2$  & 0.080240 $\pm$ 0.006853 & 0.228669 $\pm$ 0.011692 & 4.115747 $\pm$ 0.210896 &  $-3.10484 \pm 0.350603$ \\
      & CaO      & 0.959481 $\pm$ 0.052035 & 0.880215 $\pm$ 0.024368 & 7.426723 $\pm$ 0.206384 & $-13.2214 \pm 0.771261$ \\
      & Basicity & 0.008617 $\pm$ 0.000501 & 0.088803 $\pm$ 0.002844 & 4.118412 $\pm$ 0.132350 & $-11.6385 \pm 0.735167$ \\
      & Overall Mean & 10.71403 $\pm$ 6.279894
                  & 2.795390 $\pm$ 1.053220
                  & 4.612936 $\pm$ 0.184711 
                  & $-9.71642 \pm 0.620968$ \\
    \bottomrule
  \end{tabular}
\end{table*}

Following the static benchmarking, this subsection evaluates the models' online predictive performance on real-world sequential data to assess their effectiveness under nonstationary concept drift.
To ensure statistical rigor, each experiment was repeated 30 times with different random seeds on the original temporal sequence.
The detailed results across all metrics and the five quality targets are reported in Table~\ref{tab:dynamic_full}.
The results show that concept drift in real-world production leads to performance degradation for all methods compared with the shuffled setting. Nevertheless, DA-MSDL consistently outperforms the baselines across all metrics and quality targets.
In terms of the Overall Mean, DA-MSDL yields the lowest errors, suggesting that the hierarchical online adaptation mechanism helps track nonstationary dynamics and maintain reliable prediction accuracy over long horizons.
Notably, DA-MSDL shows smaller standard deviations than OB-ISSID and LSTM, indicating more stable performance under stochastic industrial disturbances and supporting more reliable inference.
The recurrent baselines (LSTM and GRU-PLS) perform poorly on long-horizon sequences, with Overall Mean MSE values exceeding 10.0, which can be attributed to limited capacity in modeling coupled multi-target dependencies and error accumulation under drift.
Although Transformer alleviates long-range dependency modeling to some extent, its fixed parameters can limit generalization when raw-material compositions fluctuate drastically.
OB-ISSID, an online identification model, shows substantial mismatch in this multi-target nonlinear task; its large variance indicates instability under strongly nonlinear operating conditions.
While Ventingformer achieves gains on some targets, it does not match the overall accuracy and stability of DA-MSDL, likely because it lacks a closed-loop online adaptation mechanism.

\begin{figure}[!t]
\centering
\includegraphics[width=\linewidth]{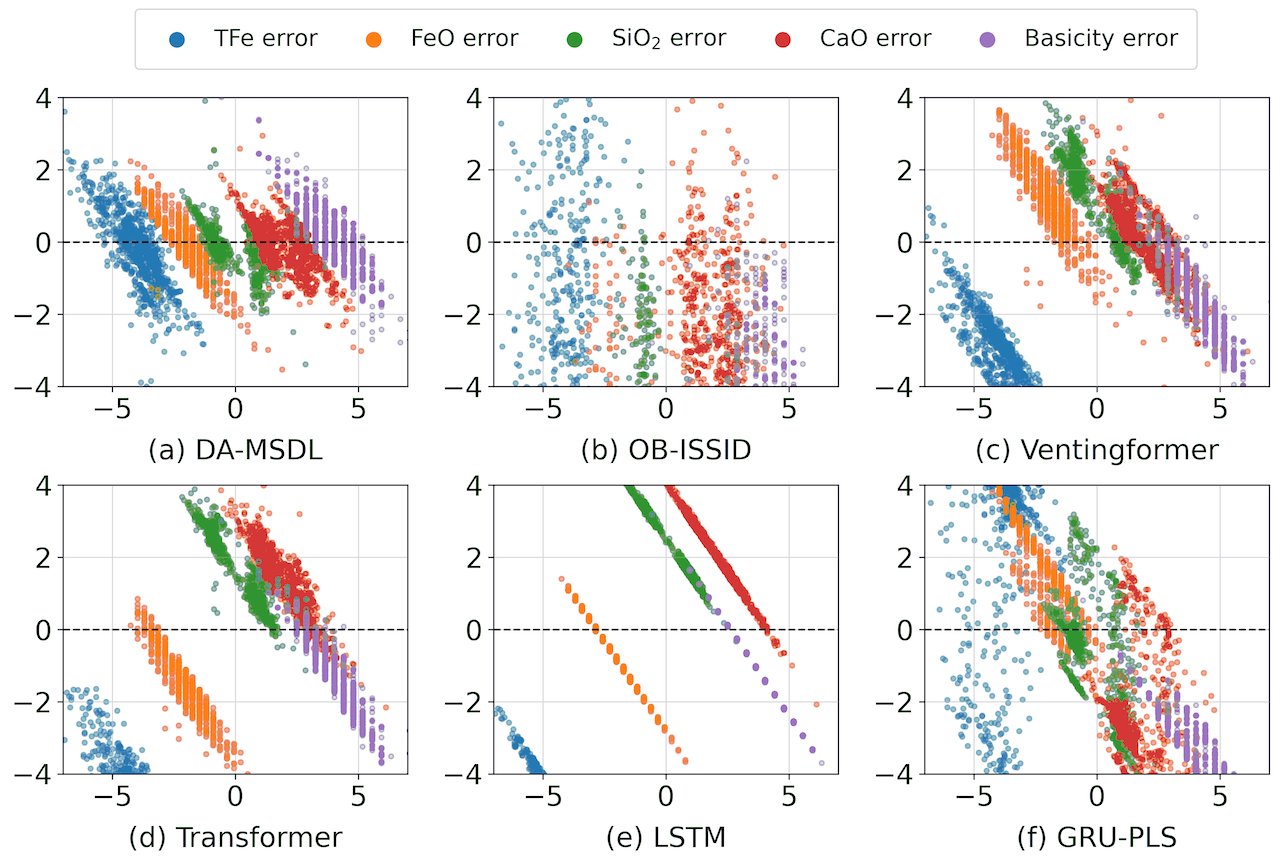}
\caption{Error scatter distribution illustrating the prediction stability across labels.}
\label{fig:dynamic_scatter}
\end{figure}

\begin{figure*}[t]
  \centering
  \includegraphics[width=\textwidth]{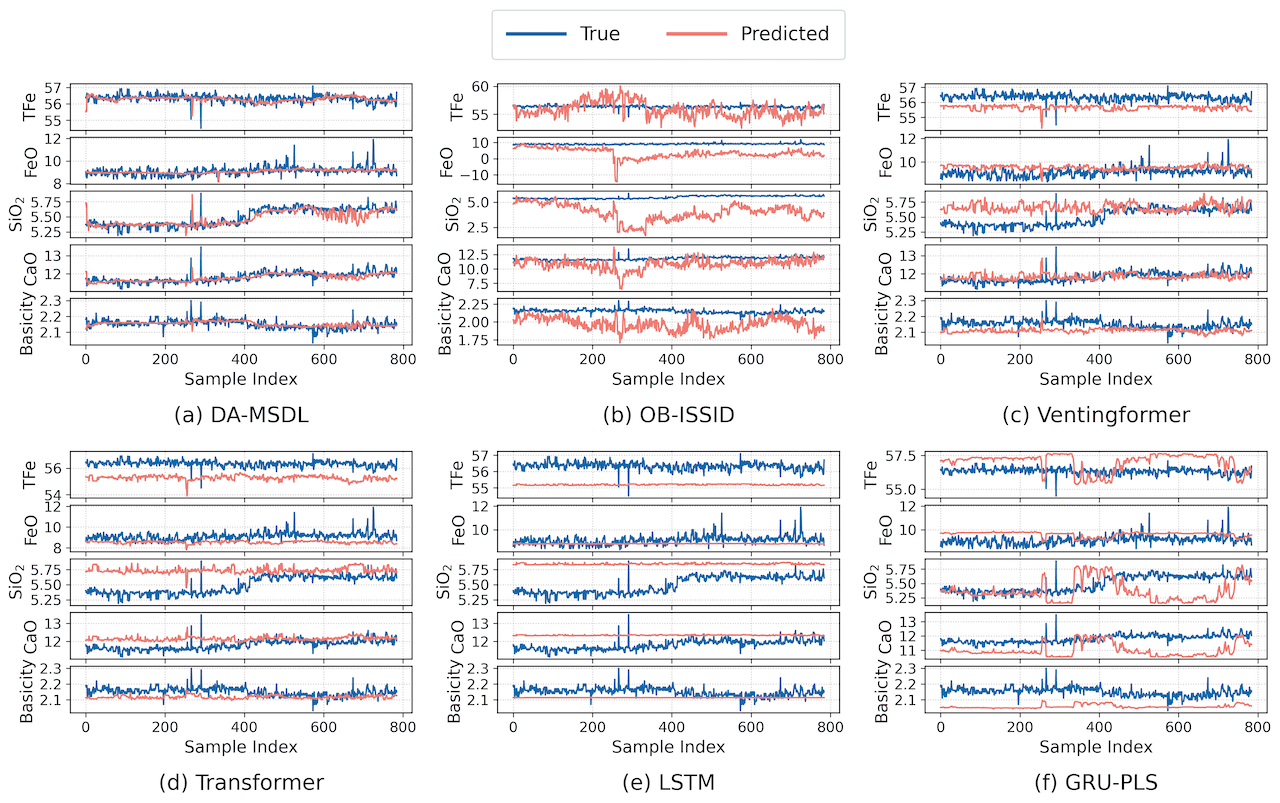}
  \caption{Time-series prediction comparison: true values vs.\ predicted values for the five quality targets on the test set across different models.}
  \label{fig:static_true_pred}
\end{figure*}

To further analyze the behavior for each target, the error scatter distributions are illustrated in Fig. \ref{fig:dynamic_scatter}.
The error point clouds of all methods exhibit a band-like structure, while DA-MSDL forms the most compact cloud: most samples cluster within a narrow band near zero error with fewer extreme outliers, indicating a more stable overall error level across the five quality targets.
While the error band of MS-BCNN is better than those of traditional baselines, it still shows a one-sided shift as errors increase; in contrast, the point clouds of OB-ISSID and GRU-PLS are more dispersed, reflecting more pronounced structural biases.
More detailed model-wise time-series comparisons are provided in Fig. \ref{fig:static_true_pred}.

\subsubsection{Generalizability on Public Industrial Benchmark}

To assess cross-domain generalizability beyond sintering, we additionally evaluate DA-MSDL on the public Water Treatment Plant dataset from the UCI Machine Learning Repository~\cite{poch1993water_treatment_plant}.
The dataset consists of multivariate process-monitoring records from an urban wastewater treatment plant and exhibits noise, missingness, and nonstationarity typical of real industrial processes.
To test generalization without dataset-specific tuning, we kept the same hyperparameter settings as those used for the sintering task.
Despite the lack of dataset-specific tuning, Table~\ref{tab:wtp_overall} shows that DA-MSDL achieves the best overall performance.
Compared with the baselines, DA-MSDL yields consistent improvements, with relative reductions of approximately 15\%--60\% on NMSE, NMAE, and MAPE.
These results indicate that the proposed framework generalizes beyond the sintering application.
This benchmark is included as supplementary evidence of cross-domain generalization; the primary focus remains real-world sintering quality prediction with delayed quality feedback.

\begin{table}[!t]
  \centering
  \caption{Overall predictive performance on the Water Treatment Plant dataset (mean $\pm$ standard deviation).}
  \label{tab:wtp_overall}
  \begin{tabular}{lccc}
    \toprule
    \textbf{Model} & \textbf{NMSE} & \textbf{NMAE} & \textbf{MAPE (\%)} \\
    \midrule
    DA-MSDL
      & \cellcolor{gray!50}{$0.7693 \pm 0.1328$}
      & \cellcolor{gray!50}{$0.5502 \pm 0.1954$}
      & \cellcolor{gray!50}{$0.55 \pm 0.03$} \\
    OB-ISSID
      & $1.3479 \pm 0.4521$
      & $0.6145 \pm 0.2545$
      & $0.63 \pm 0.04$ \\
    Ventingformer
      & $1.7582 \pm 0.9545$
      & $0.8336 \pm 0.7451$
      & $0.85 \pm 0.10$ \\
    Transformer
      & $4.3388 \pm 1.5265$
      & $1.6894 \pm 1.6541$
      & $1.71 \pm 0.24$ \\
    LSTM
      & $0.9525 \pm 0.4536$
      & $0.6247 \pm 0.2654$
      & $0.64 \pm 0.04$ \\
    GRU-PLS
      & $1.3018 \pm 0.5562$
      & $0.7286 \pm 0.2684$
      & $0.74 \pm 0.05$ \\
    \bottomrule
  \end{tabular}
\end{table}

\subsection{Ablation Study}

\subsubsection{Effectiveness of the Dynamic Adaptation Mechanism}

To assess the contribution of dynamic adaptation under drift, we compare DA-MSDL against the static backbone MS-BCNN in online prediction.

\begin{figure}[!t]
  \centering
  \includegraphics[width=\linewidth]{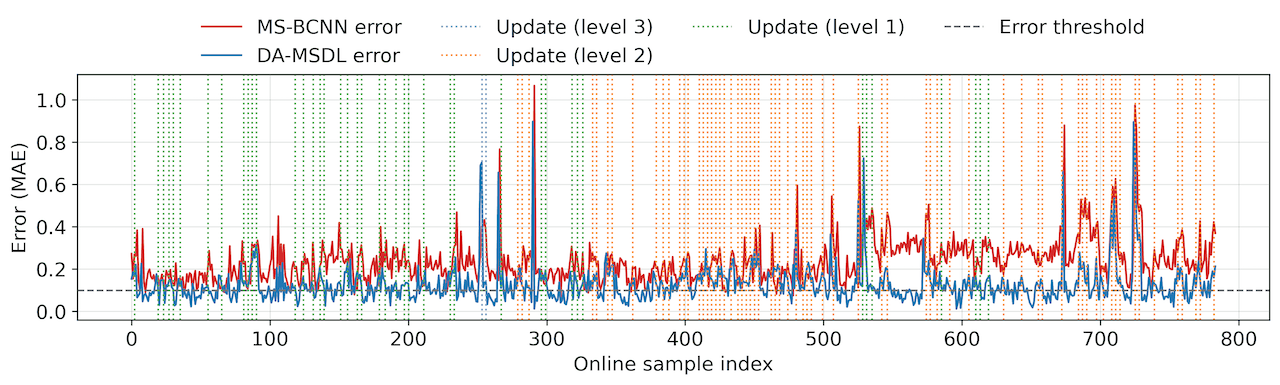} 
  \caption{Online MAE of MS-BCNN (static) and DA-MSDL (dynamic) on real-world production sequences. Dashed lines mark detected concept drift events where online adaptation is triggered.}
  \label{fig:drift_compare}
\end{figure}
Fig.~\ref{fig:drift_compare} compares the online MAE of the two methods and highlights the detected drift events.
In the early stage and under approximately steady conditions, the two methods achieve comparable accuracy.
As drift accumulates, MS-BCNN shows a rising error trend and pronounced spikes around several detected drift points (dashed lines).
By contrast, DA-MSDL triggers online fine-tuning after drift detection, allowing the error to return quickly to a stable range and preventing prolonged degradation.

\begin{figure}[!t]
  \centering
  \includegraphics[width=\linewidth]{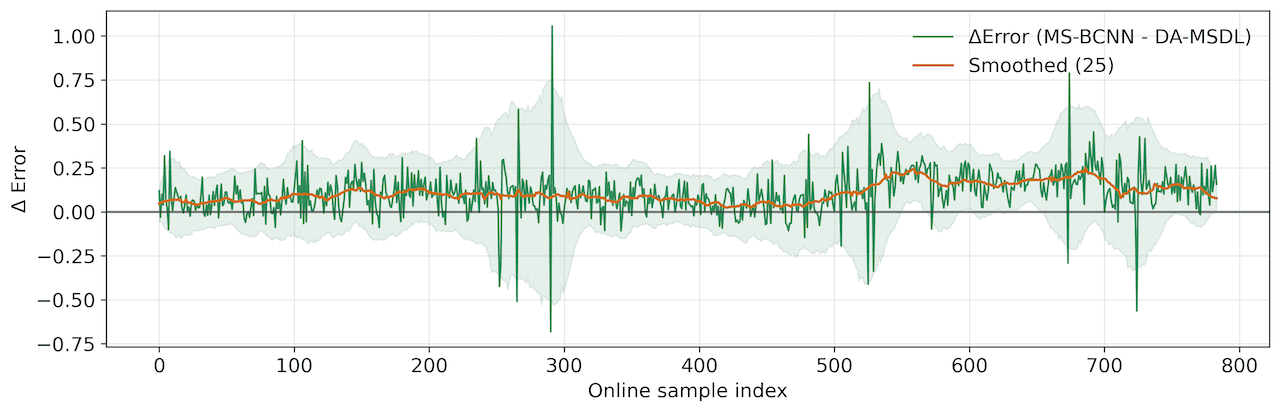} 
  \caption{Error difference ($\Delta Error$) between MS-BCNN and DA-MSDL over time. Persistent positive values indicate consistent gains from dynamic adaptation.}
  \label{fig:error_trend}
\end{figure}
To quantify long-term gains, Fig.~\ref{fig:error_trend} plots the error difference ($\Delta Error$) between MS-BCNN and DA-MSDL over the online sequence.
$\Delta Error$ stays mostly above zero throughout the sequence, indicating a persistent advantage of DA-MSDL over the static backbone; the gap widens during periods of stronger operating-condition fluctuations.

\subsubsection{Component-wise Contribution Analysis}

\begin{figure}[!t]
  \centering
  \begin{minipage}{0.48\linewidth}
    \centering
    \includegraphics[width=\linewidth]{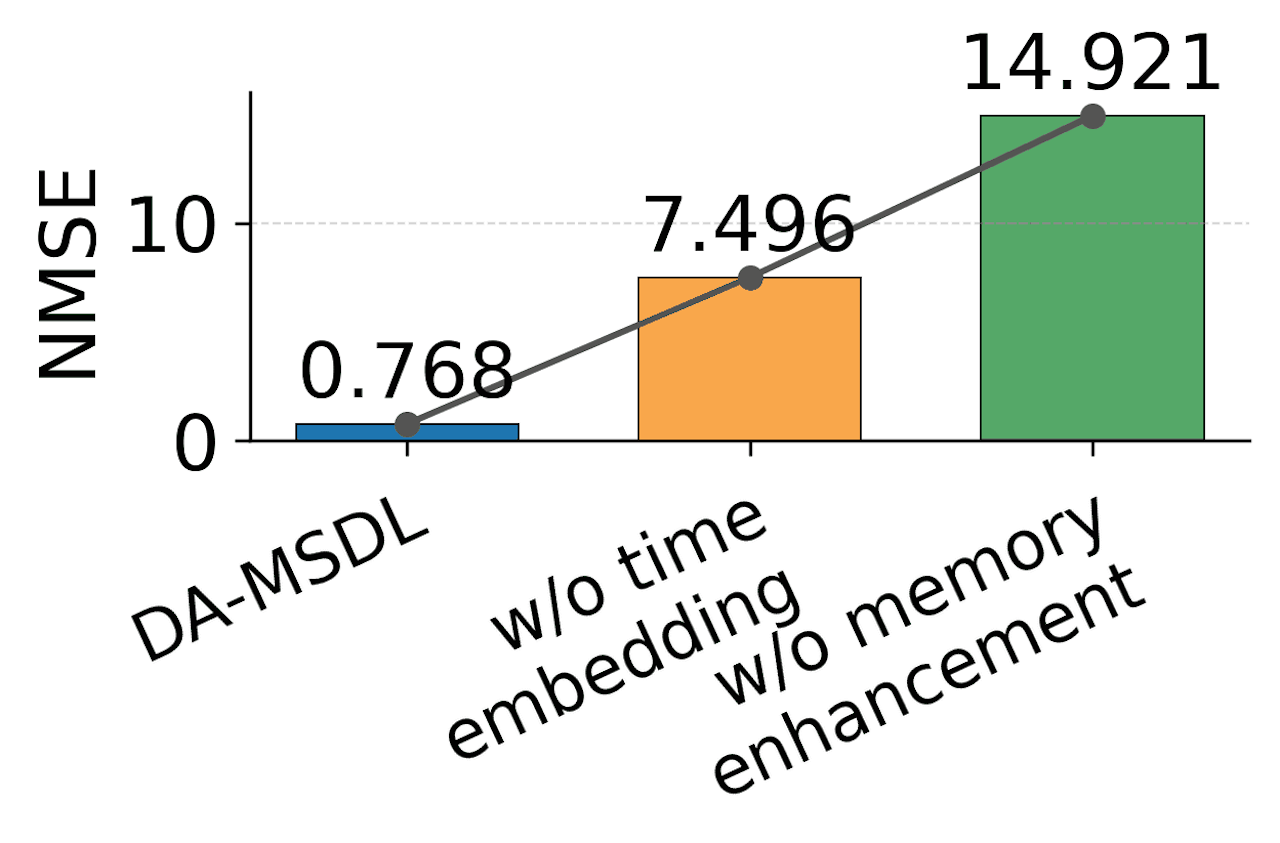}
    \\[0.2em]
    \small (a) Multi-scale sampling and memory enhancement
  \end{minipage}
  \hfill
  \begin{minipage}{0.48\linewidth}
    \centering
    \includegraphics[width=\linewidth]{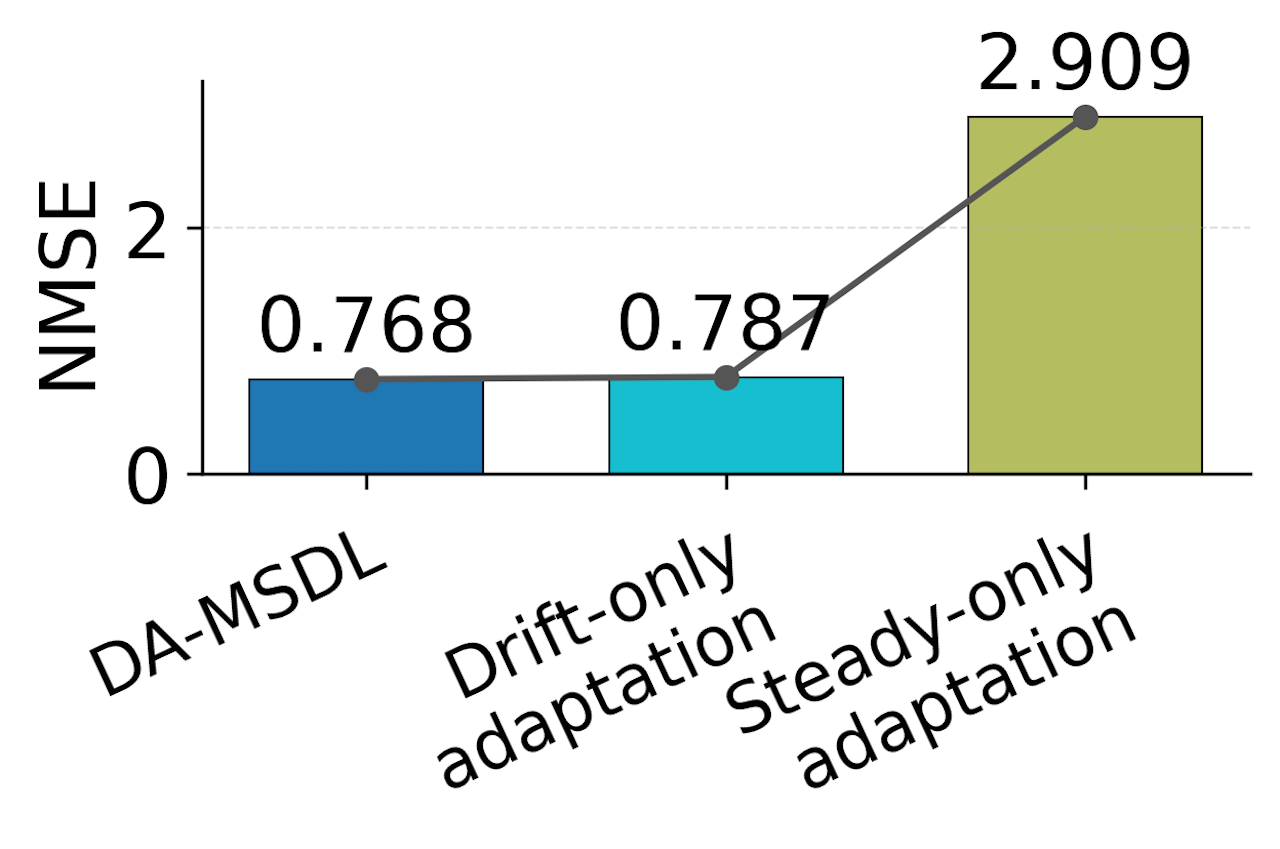}
    \\[0.2em]
    \small (b) Drift-triggered vs.\ steady-state adaptation
  \end{minipage}

  \vspace{0.4em}

  \begin{minipage}{0.48\linewidth}
    \centering
    \includegraphics[width=\linewidth]{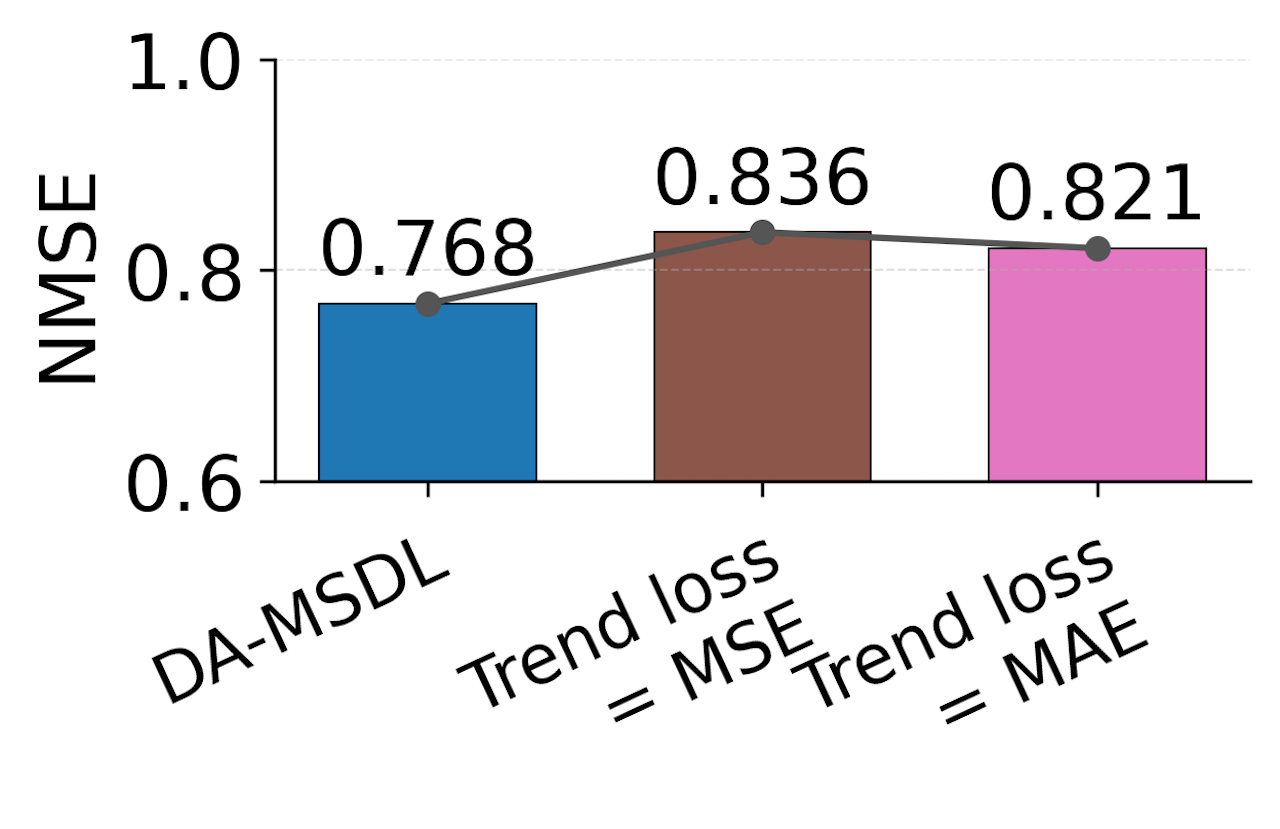}
    \\[0.2em]
    \small (c) Trend-aware loss variants
  \end{minipage}
  \hfill
  \begin{minipage}{0.48\linewidth}
    \centering
    \includegraphics[width=\linewidth]{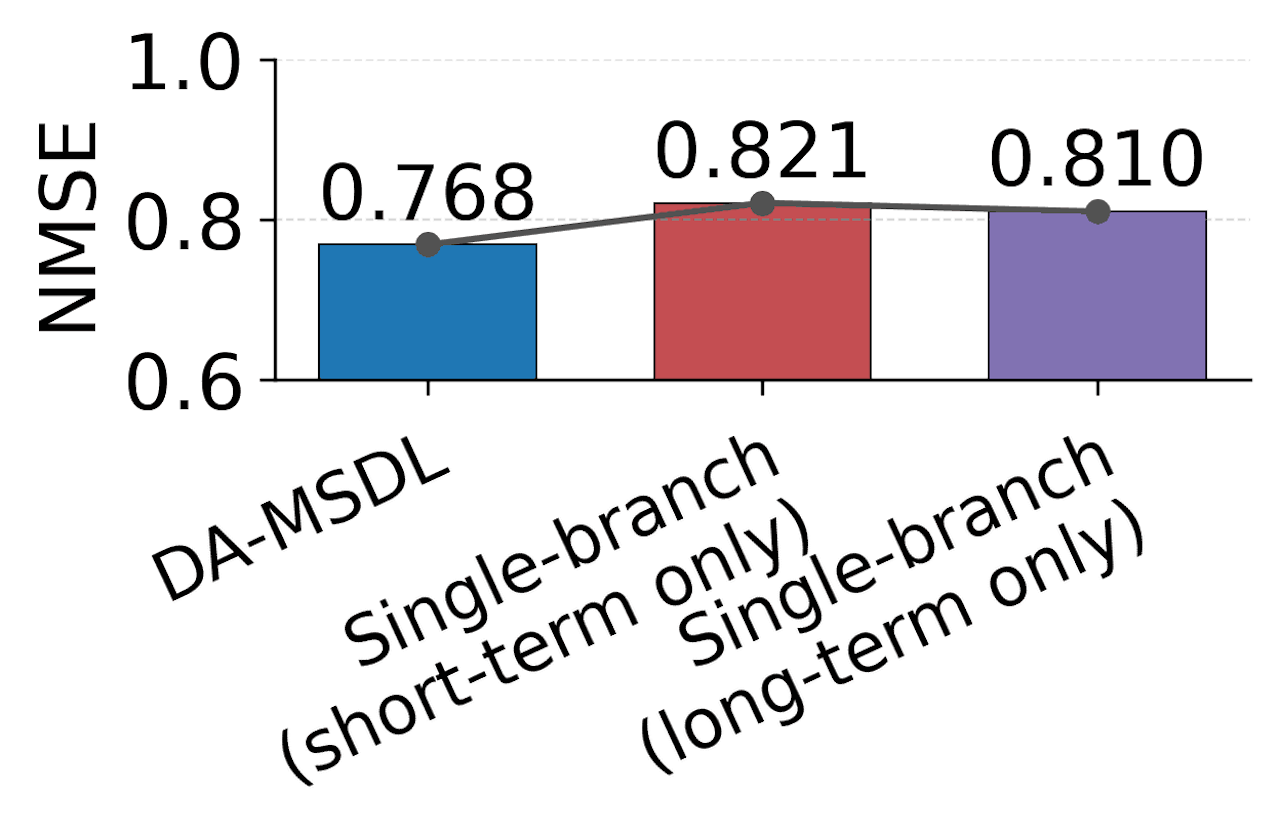}
    \\[0.2em]
    \small (d) Dual-branch convolution design
  \end{minipage}

  \caption{Component-wise ablation of DA-MSDL measured by overall NMSE. The results show how each component affects prediction performance under nonstationary conditions.}
  \label{fig:ablation_overall}
\end{figure}

To quantify component-wise contributions, we evaluate multiple DA-MSDL variants on the real-world industrial dataset, with overall results summarized in Fig.~\ref{fig:ablation_overall}.
As shown in Fig.~\ref{fig:ablation_overall}(a), time embedding and the memory module are key contributors to overall performance.
Removing time embedding substantially increases NMSE, whereas removing the memory module causes a pronounced degradation in predictive accuracy.
This suggests that sliding windows alone are insufficient to maintain long-term temporal dependencies under nonstationarity, and that the memory mechanism is important for retaining information across drift periods.
Fig.~\ref{fig:ablation_overall}(b) compares variants with different adaptation strategies.
The ``drift-only'' variant achieves accuracy close to the full model, whereas the ``steady-only'' variant (without drift response) shows a clear increase in prediction error.
This indicates that drift-triggered response is the main driver of robustness, while steady-state updates provide incremental refinements.
Regarding the objective design (Fig.~\ref{fig:ablation_overall}(c)), replacing the trend-aware loss with MSE or MAE degrades predictive performance.
By penalizing trend deviations, the model better captures gradual, evolving changes in quality targets beyond point-wise fitting.
Finally, the full dual-branch architecture outperforms both single-branch variants (Fig.~\ref{fig:ablation_overall}(d)). 
These results suggest that jointly modeling short-term fluctuations and long-term trends is important for capturing coupled multi-target dependencies in complex industrial processes.

\subsubsection{Drift Mechanism Analysis}

To evaluate the responsiveness of drift detection and hierarchical fine-tuning, we measure how many online samples are needed for the error to return to a stable range after each adaptation trigger.

\begin{figure}[!t]
\centering
\includegraphics[width=0.7\linewidth]{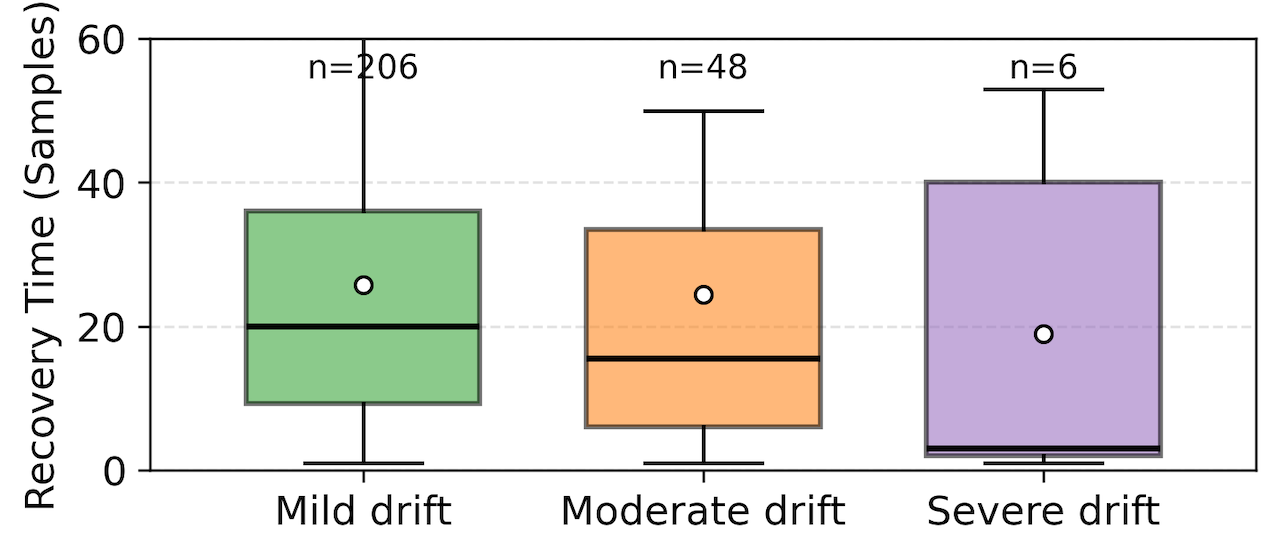}
\caption{Recovery-time distribution under different drift severity levels.}
\label{fig:recovery_time}
\end{figure}
As shown in Fig.~\ref{fig:recovery_time}, ``recovery time'' is defined as the number of online samples from drift detection (and the corresponding fine-tuning trigger) until the online MAE drops below the threshold of the associated severity level.
The results show that mild drift events recover quickly, with most cases returning below the threshold within a small number of samples.
For moderate drifts, recovery remains comparable to mild cases, consistent with the higher fine-tuning intensity assigned by the hierarchical strategy.
Severe drift events are less frequent, but they trigger higher-intensity updates under the hierarchical strategy; their recovery times nevertheless exhibit a broader distribution, consistent with the limited number of observed occurrences.
Overall, recovery is achieved within a short span across severity levels, suggesting that the assigned fine-tuning intensities are well matched to the detected drift severity.

Consistent with the error trajectories in Fig.~\ref{fig:drift_compare}, DA-MSDL reduces error rapidly after each trigger and then remains stable, whereas MS-BCNN stays at elevated error levels without an adaptation mechanism.
These observations suggest that hierarchical fine-tuning is more than additional computation: it provides targeted updates that help the model track the evolving data distribution under drift.

\subsubsection{Sensitivity Analysis of Key Hyperparameters}

To assess sensitivity to key hyperparameters, we analyze the effects of the memory queue length $T_m$, the trend-supervision horizon $H$, and the drift detection window length $L_w$.
\begin{figure}[t]
  \centering
  \begin{minipage}{0.32\linewidth}
    \centering
    \includegraphics[width=\linewidth]{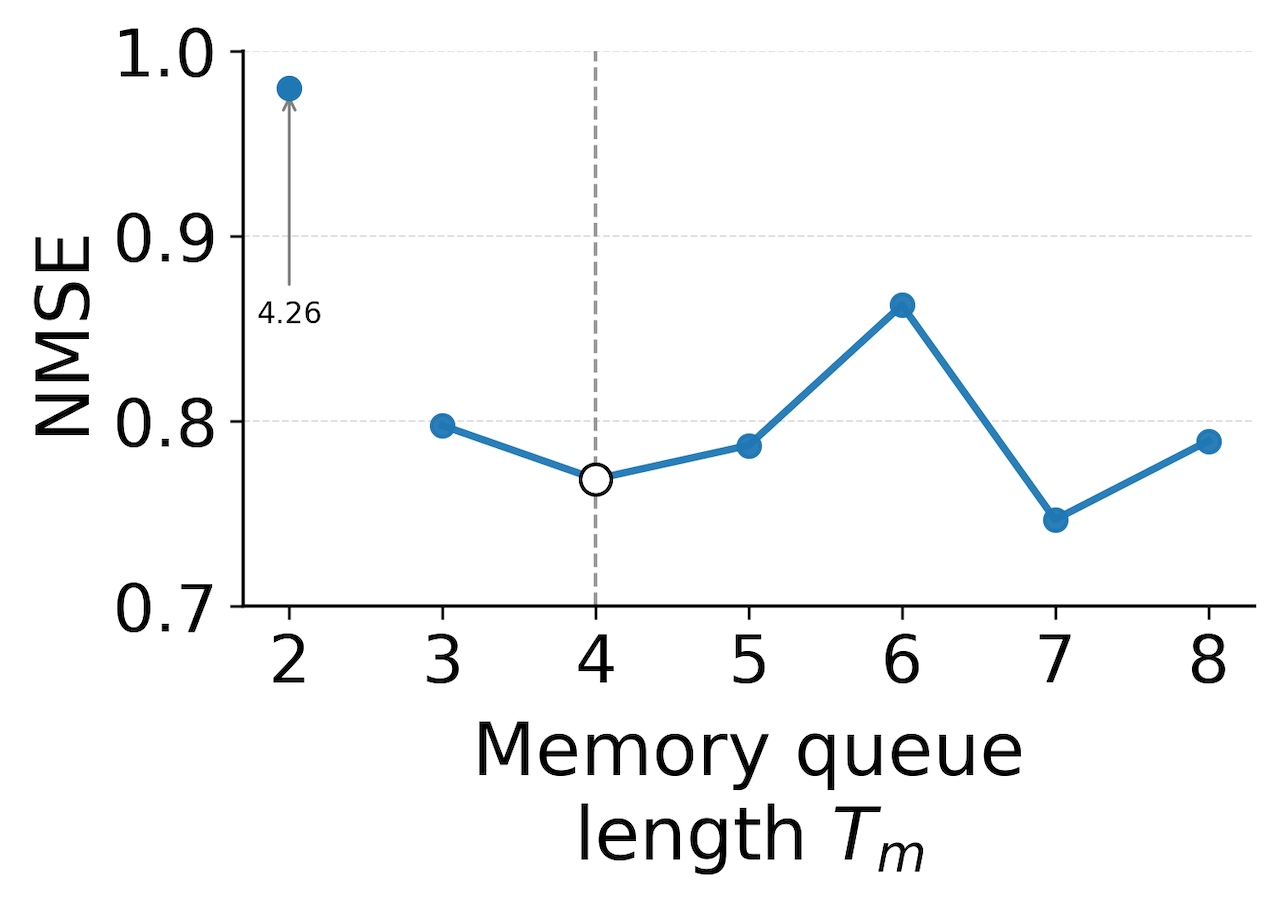}
    \\[0.2em]
    \small (a) Sensitivity to memory queue length
  \end{minipage}
  \hfill
  \begin{minipage}{0.32\linewidth}
    \centering
    \includegraphics[width=\linewidth]{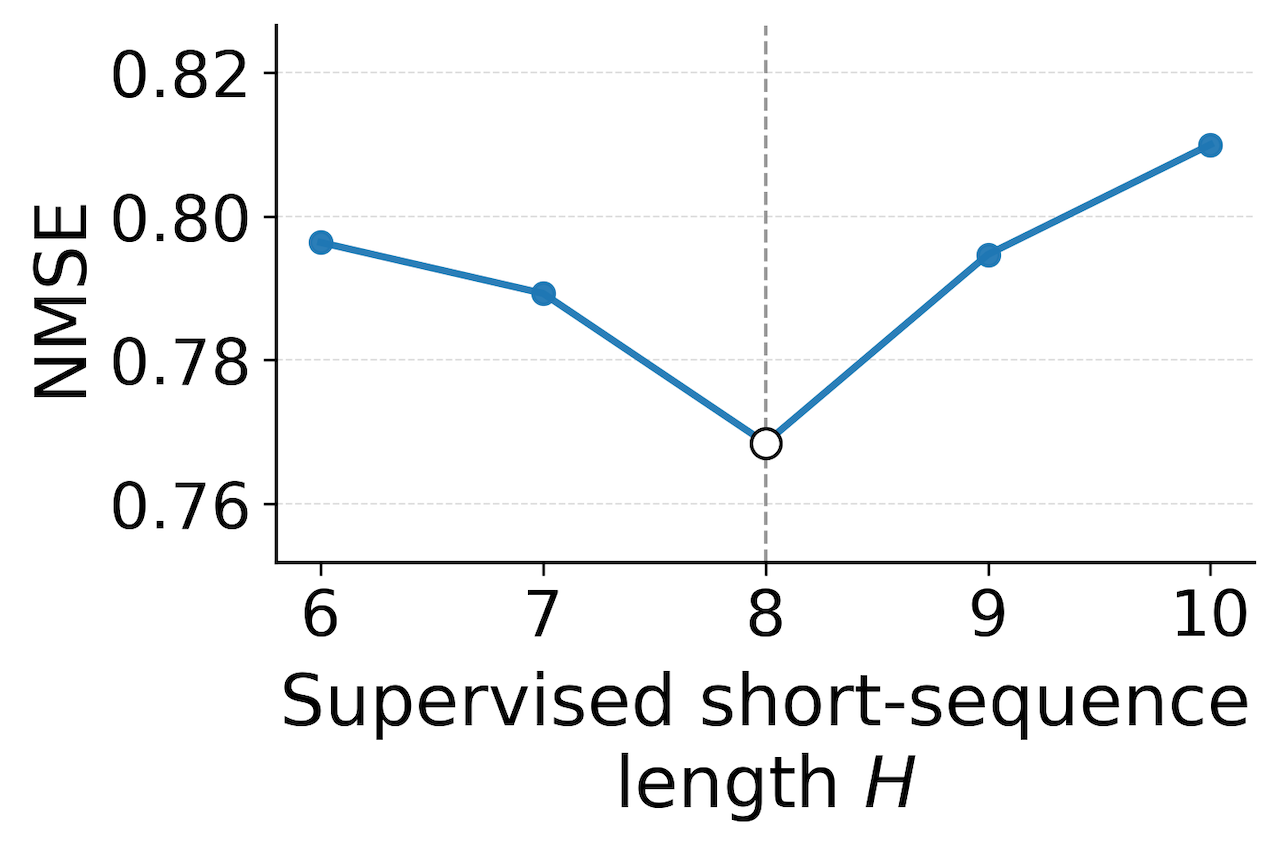}
    \\[0.2em]
    \small (b) Sensitivity to supervised short-sequence length
  \end{minipage}
  \hfill
  \begin{minipage}{0.32\linewidth}
    \centering
    \includegraphics[width=\linewidth]{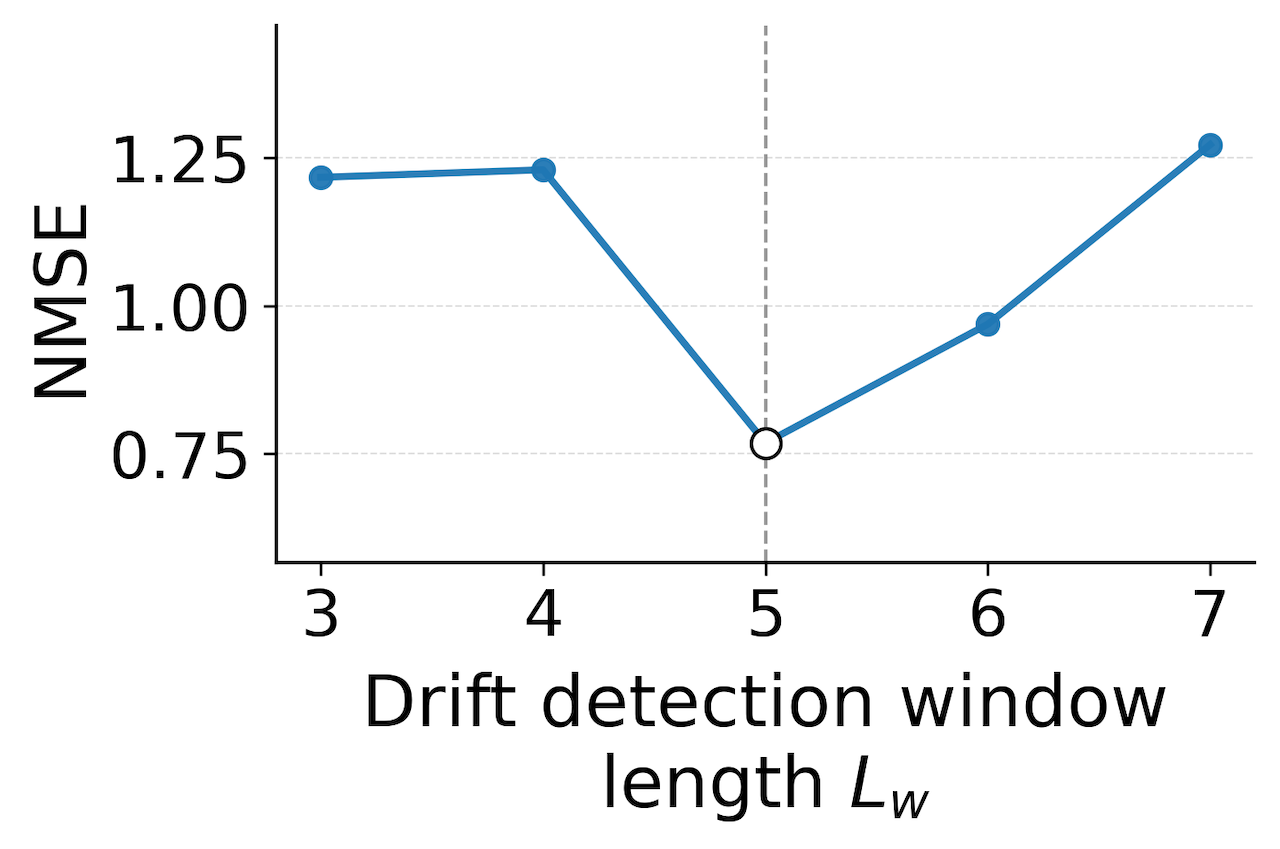}
    \\[0.2em]
    \small (c) Sensitivity to drift detection window length
  \end{minipage}
  \caption{Sensitivity analysis of DA-MSDL with respect to key hyperparameters.}
  \label{fig:param_sensitivity}
\end{figure}
The effect of each hyperparameter is evaluated via univariate experiments experiments with all other settings fixed; the results are summarized in Fig.~\ref{fig:param_sensitivity}.
As shown in Fig.~\ref{fig:param_sensitivity}(a), performance degrades when the memory queue is too short, suggesting that limited historical information is insufficient to regularize learning under nonstationarity.
As $T_m$ increases, NMSE improves and then stabilizes on a plateau, indicating stable performance within a reasonable look-back range.
Balancing accuracy and computational cost, we set $T_m=4$ as the default within the plateau region.
Fig.~\ref{fig:param_sensitivity}(b) shows the effect of the trend-supervision horizon $H$, which exhibits a U-shaped pattern.
A moderate choice (e.g., $H=8$) balances smoothing local fluctuations and preserving trend responsiveness, whereas overly small or large $H$ degrades the modeling of gradual evolution.
In contrast, $L_w$ is more sensitive (Fig.~\ref{fig:param_sensitivity}(c)): a small window may trigger adaptations too frequently, whereas a large window delays responses to operating-condition drift.
Under the current operating conditions, $L_w=5$ provides a good trade-off by filtering random perturbations while remaining responsive to underlying distribution shifts.
Overall, DA-MSDL remains stable over reasonable hyperparameter ranges, indicating robustness to configuration variations.

\section{Conclusion and Future Work}
\label{Section5}

This paper proposes a drift-aware online dynamic learning framework, DA-MSDL, to address pronounced nonstationarity and severe delayed quality feedback in industrial sintering.
DA-MSDL identifies distribution shifts without immediate labels via unsupervised MMD-based drift detection and adopts a proactive ``detect--fine-tune--predict'' pipeline for feed-forward online adaptation.
Moreover, a drift-severity-guided hierarchical fine-tuning strategy, together with a dynamic memory queue and prioritized replay, enables a controlled stability--plasticity trade-off and improves long-horizon predictive reliability.
Experiments on a real-world industrial sintering data stream and a public industrial benchmark show that DA-MSDL consistently outperforms representative baselines under concept drift and delayed supervision, while ablation and sensitivity studies support the effectiveness and robustness of the key modules.

While DA-MSDL yields stable gains on long-horizon industrial data streams, several directions merit further investigation.
First, we will investigate integrating process physics into online learning (e.g., physics-consistent constraints or hybrid modeling) to improve interpretability and extrapolation under extreme conditions and label-scarce periods.
Second, we will develop model compression and resource-aware online updating for edge deployment, reducing fine-tuning cost while maintaining reliable drift responsiveness toward real-time closed-loop applications.

\IEEEpubidadjcol

\bibliographystyle{IEEEtran}
\bibliography{refs}

\end{document}